\pdfoutput=1

\documentclass[11pt]{article}

\usepackage[final]{EMNLP2023}

\usepackage{times}
\usepackage{latexsym}

\usepackage[T1]{fontenc}

\usepackage[utf8]{inputenc}

\usepackage{microtype}

\usepackage{inconsolata}

\usepackage{xspace,mfirstuc,tabulary}
\usepackage{booktabs}
\usepackage{amssymb}
\usepackage{amsmath}
\usepackage{multirow,booktabs, hhline}
\usepackage[ruled,noend]{algorithm2e}
\usepackage{amsmath, bm}
\usepackage{graphicx}
\usepackage{cleveref}
\usepackage{bm}
\usepackage{mathtools,tikz,caption}
  
\crefname{section}{§}{§§}

\usepackage{makecell}
\usepackage{adjustbox}
\usepackage{xcolor}
\usepackage{color, colortbl}

\definecolor{BLUE_D1}{HTML}{a9cce3}
\definecolor{ORANGE_D1}{HTML}{f5cba7}
\definecolor{GREEN_D1}{HTML}{b6d7a8}
\definecolor{GRAY_D1}{HTML}{d5dbdb}
\usepackage{graphicx}

\usepackage{bbm}
\usepackage{float}
\usepackage{subfigure}
\usepackage{warpcol}
\usepackage{makecell}
\usepackage{hyperref}
\usepackage{dcolumn,booktabs}

\newcommand\PET{PET\xspace}
\newcommand\APET{APET\xspace}
\newcommand\APETADJACENT{APET$_{\textsc{Adjacent}}$\xspace}
\newcommand\APETDISCRETE{APET$_{\textsc{Discrete}}$\xspace}

\newcommand\TSMALL{T5$_{\textsc{small}}$\xspace}
\newcommand\TBASE{T5$_{\textsc{base}}$\xspace}

\newcommand\TXXL{T5$_{\textsc{xxl}}$\xspace}

\newcommand\BERTSMALL{BERT$_{\textsc{small}}$\xspace}
\newcommand\BERTBASE{BERT$_{\textsc{base}}$\xspace}
\newcommand\BERTLARGE{BERT$_{\textsc{large}}$\xspace}

\newcommand\BLOOMSMALL{BLOOM$_{\textsc{560M}}$\xspace}
\newcommand\BLOOMBASE{BLOOM$_{\textsc{1.1B}}$\xspace}
\newcommand\BLOOMLARGE{BLOOM$_{\textsc{7.1B}}$\xspace}

\newcommand{\figref}[1]{Figure \ref{#1}}

\definecolor{b5de2a}{RGB}{181, 222, 42}
\definecolor{gr}{rgb}{0.0, 0.5, 0.0}
\definecolor{gr_light}{RGB}{169,209,142}
\definecolor{def_deep_blue}{RGB}{0,0,255}
\definecolor{def_deep_blue_light}{RGB}{169,209,142}
\definecolor{def_deep_green}{RGB}{44,160,44}
\definecolor{def_deep_purple}{RGB}{191,0,191}
\definecolor{def_deep_orange}{RGB}{249,115,8}
\definecolor{def_deep_orange_light}{RGB}{244,178,132}
\definecolor{def_light_blue}{RGB}{1,191,255}
\definecolor{def_light_green}{RGB}{50,205,50}
\definecolor{def_light_purple}{RGB}{255,105,180}
\definecolor{def_light_orange}{RGB}{255,165,0}

\newcolumntype{d}[1]{D{.}{.}{#1}}

\newcommand\mctwo[1]{\multicolumn{2}{c}{#1}} 
\newcommand\mctwol[1]{\multicolumn{2}{c|}{#1}}

\title{Exploring the Impact of Model Scaling on Parameter-efficient Tuning}

\author{
 Yusheng~Su$^{1}\thanks{\quad The first two authors contributed equally.}$\hspace{0.5em}, Chi-Min~Chan$^{1*}$, Jiali Cheng$^2$, Yujia~Qin$^{1}$, Yankai~Lin$^3$, \textbf{Shengding Hu}$^1$,\\ \textbf{Zonghan Yang}$^1$, \textbf{Ning Ding}$^1$, \textbf{Xingzhi Sun}$^4$, \textbf{Guotong Xie}$^4$, \textbf{Zhiyuan Liu}$^1$\thanks{\quad Corresponding author: Z.Liu and M.Sun.}, \textbf{Maosong Sun}$^{1+}$ \\
 $^1$Department of Computer Science and Technology, Tsinghua University \\
 $^2$University of Massachusetts Lowell \\
 $^3$Gaoling School of Artificial Intelligence, Renmin University \\
 $^4$Ping An Technology \\
    \tt{yushengsu.thu@gmail.com}\\
}

\begin{document}
\maketitle

\begin{abstract}
Parameter-efficient tuning (\PET) methods can effectively drive extremely large pre-trained language models (PLMs) by training only minimal parameters. Different \PET methods utilize different manually designed tunable modules. In small PLMs, there are usually noticeable performance differences among \PET methods. Nevertheless, as the model scale increases, the performance differences become marginal. Hence, we hypothesize that model scaling mitigates the impact of design differences on \PET methods. To investigate this hypothesis, we introduce a more flexible \PET method called Arbitrary \PET (\APET) method. The \APET method is compatible with a tunable module, which consists of any number of parameters distributed in arbitrary positions. Then, we utilize it and conduct experiments on $11$ NLP tasks across $3$ representative PLMs. Our investigations reveal that model scaling (1) mitigates the effects of the positions of tunable parameters on performance, and (2) enables tuning methods to achieve performance comparable to full-parameter fine-tuning by optimizing fewer tunable parameters. Intriguingly, we also observe that tuning methods optimize the similar number of tunable parameters to exceed random guess performance on different tasks. We collectively discuss this phenomenon and the two aforementioned findings from an optimization perspective to understand the underlying mechanisms. These conclusions enhance our understanding of the impact of model scaling on \PET and assist in designing more effective and efficient \PET methods for PLMs of different scales. The source code can be obtained from this GitHub repository: \url{https://github.com/yushengsu-thu/PET_Scaling}.
\end{abstract}

\begin{figure}[!t]
\centering
\includegraphics[width=0.49\textwidth]{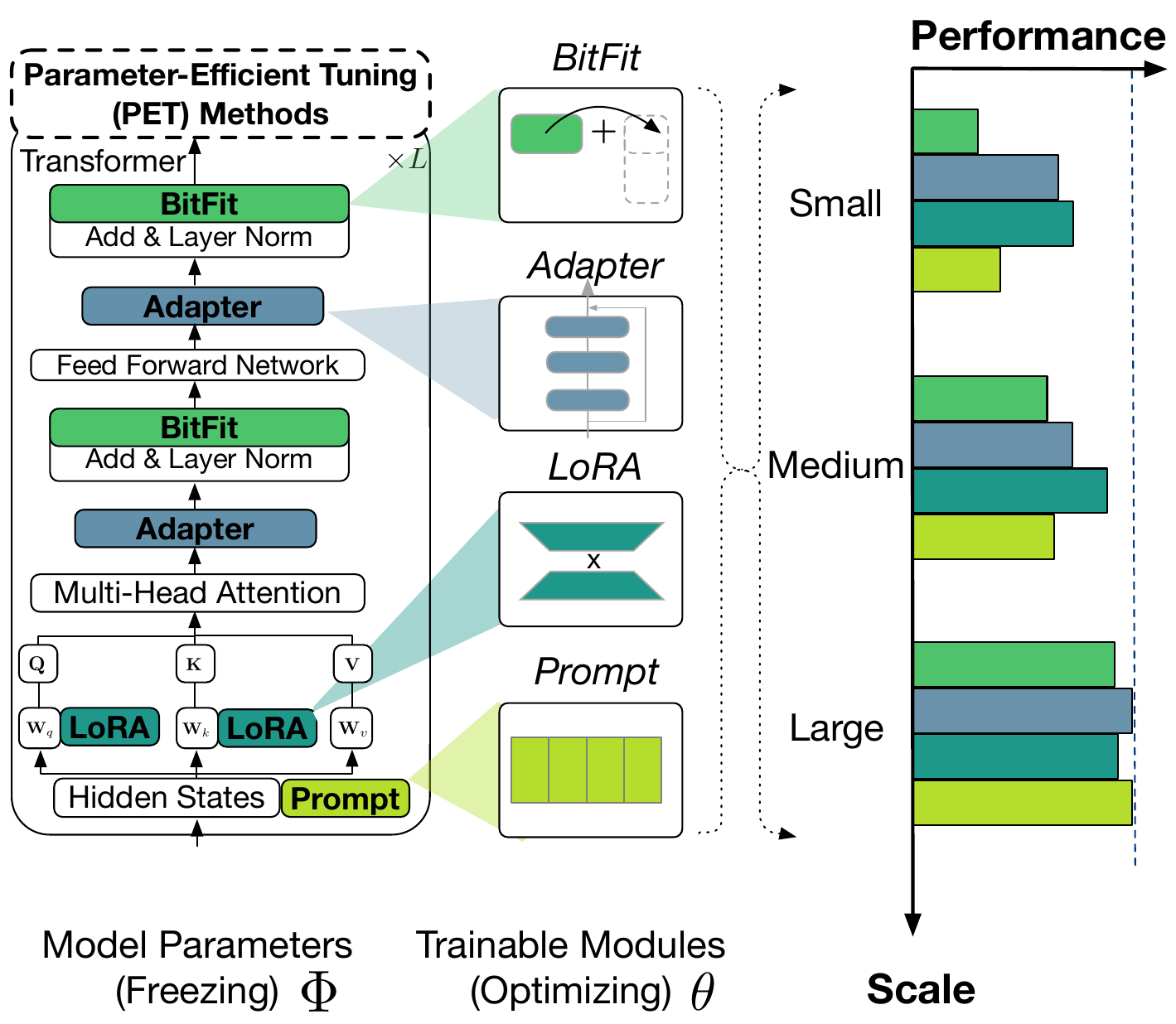}
\caption{Different \PET methods have distinct tunable modules, which typically result in noticeable performance differences. However, as the model scale increases, these differences become less significant.}
\label{fig:the_power_of_scale}
\end{figure}

\section{Introduction}
Pre-trained language models (PLMs), such as GPT \cite{radford2018improving}, BERT \cite{devlin2018bert}, and T5 \cite{raffel2020exploring}, have achieved great success on various natural language processing (NLP) tasks. Despite their effectiveness, fine-tuning (FT) these large-scale PLMs with full parameters incurs both unaffordable computational and storage costs. To solve this problem, researchers have proposed a series of parameter-efficient tuning (\PET) methods \cite{pmlr-v97-houlsby19a,li-liang-2021-prefix,NEURIPS2021_081be9fd,lester-etal-2021-power,compacter2021,hu2022lora,ben-zaken-etal-2022-bitfit,sparseadapter2022} which only update an assigned tunable module consisting of minimal parameters while freezing the rest parameters in a PLM during model adaptation.

Although these existing representative \PET methods can reduce computational and storage costs, there are usually noticeable performance differences among these representative \PET methods on downstream tasks. Intriguingly, as the scale of a PLM increases, the performance differences among \PET methods become narrower, as illustrated in Figure \ref{fig:the_power_of_scale}. These findings are interesting and worth exploring because the existing representative \PET methods are designed with disparate philosophies, e.g., tunable modules that are composed of \textbf{different numbers} of tunable parameters distributed in \textbf{arbitrary positions}. Hence, we hypothesize that \textit{model scaling mitigates the effects of the above design differences among the \PET methods on performance}. To validate this hypothesis, we further conduct two lines of ablation analyses: 
\begin{enumerate}
\itemsep=-4pt
\item[(A1)] Whether the model scale mitigates the performance differences resulting from the position of tunable parameters.
\item[(A2)] Whether the model scale mitigates the performance differences resulting from the number of tunable parameters.
\end{enumerate}
However, solely investigating the four representative \PET methods (see Figure \ref{fig:the_power_of_scale}) might be insufficient to encompass an adequate range of parameter positions for the ablation analyses (A1). Additionally, the tunable modules of these four \PET methods are constrained to be composed of layer-level tensors or matrices, making it challenging to precisely control the number of tunable parameters at the fine-grained (parameter) level in the ablation analyses (A2). To facilitate the ablation analyses, we develop a more flexible \textbf{A}rbitrary \textbf{P}arameter-\textbf{E}fficient \textbf{T}uning (\APET) method (\cref{ssec:afp_tuning_methods}), which can be compatible with any number of tunable parameters distributed in arbitrary positions.

In analysis (A1), we compare the performance of \APET methods with an equal number of tunable parameters distributed in different positions. Based on the experimental results, we observe smaller differences in the performance of these \APET methods on larger models. This finding suggests that scaling the model \textit{mitigates the effects caused by the position of tunable parameters on performance}.

In analysis (A2), we compare the performance of the same \APET methods with varying numbers of tunable parameters. Based on the experimental results, we observe that model scaling does not mitigate the effects caused by the number of tunable parameters on performance. Furthermore, we have observed two interesting phenomena when the number of tunable parameters reaches two thresholds: the high threshold and the low threshold. When the number of tunable parameters equals the high threshold, \APET methods can achieve the full-parameter fine-tuning performance of the corresponding backbone model, and the high threshold tends to be lower on the larger models. Namely, \textit{\PET methods can optimize fewer tunable parameters to achieve full-parameter fine-tuning performance on the larger models}. On the other hand, when the number of tunable parameters exceeds the low parameter threshold, all \APET methods outperform random guess performance. We find that the low thresholds are nearly identical across the same models, even for different tasks. This suggests that \textit{across different tasks, \PET methods can optimize a similar number of tunable parameters on the same PLM to surpass random guess performance}.

In summary, we introduce a more flexible \PET methods - \APET methods - to conduct the extensive ablation analyses and reveal the impact of model scaling on \PET design, e.g., (1) the position of tunable parameters (\cref{ssec:ms_model_structure}) and (2) the number of tunable parameters (\cref{ssec:ms_trainable_parameter}). (3) Furthermore, we discuss the findings of ablation analyses from the perspective of optimization (\cref{sec:discussion}). We hope these conclusions not only encourage more researchers to explore the impact of model scaling on tuning methods from a theoretical perspective, but also provide guidance for designing tuning methods for models of different scales.


\section{Related Work}
\label{label:related_work}
\paragraph{Parameter-Efficient Tuning (\PET) Methods}
\label{par:pet}
With larger PLMs continuously being developed, fine-tuning all of the parameters and storing the adapted weights become increasingly cumbersome. To address the issue, researchers propose \PET methods which keep most of the parameters of PLMs frozen and optimize only a tunable module consisting of a few parameters during downstream adaptation. Over the recent years, many different designs of \PET methods have emerged. For instance, some \PET methods insert the external tunable modules after the feed-forward and attention layers in a PLM \citep{pmlr-v97-houlsby19a,pfeiffer-etal-2021-adapterfusion,karimi2021compacter}; others prepend the tunable modules into attention layers \citep{li-liang-2021-prefix,hu2022lora} or the embedding layer \citep{lester-etal-2021-power}. Another line of \PET method selects the existing parameters in a PLM \citep{ben-zaken-etal-2022-bitfit,guo2020parameter} as the tunable module to optimize. To further enhance the performance of \PET methods, some works propose automatic selection strategies \cite{hu2022sparse,chen2023parameterefficient,lawton-etal-2023-neural,zhou2023autopeft} for tunable parameters.

\begin{table*}[!t]
\begin{center} 
\begin{adjustbox}{max width=0.98\linewidth}
{
\setlength\tabcolsep{0.25em}
\begin{tabular}{cc|cc|cc}
\toprule
\multicolumn{2}{c|}{\multirow{1}{*}{\textbf{\PET Methods}}} & \multicolumn{2}{c|}{Unified View of \PET Methods} & \multicolumn{2}{c}{Positions of Tunable Modules $\theta=\{\mathbf{W}_{1}, \mathbf{W}_{2},...,\mathbf{W}_{p}\}$}\\ 
\midrule

\multicolumn{2}{c|}{Prompt \cite{lester-etal-2021-power}} & \mctwol{\multirow{4}{*}{$\mathbf{h}^{out} = f(\mathbf{h}^{in})+\Delta \mathbf{h}$}} & \mctwo{\textbf{W} will be \textbf{concatenated} to \textbf{input hidden states}} \\

\multicolumn{2}{c|}{Adapter \cite{pmlr-v97-houlsby19a}} & \mctwol{} & \mctwo{\textbf{W} will be \textbf{plugged} between \textbf{SelfAttn./FFN. layers}}\\
\multicolumn{2}{c|}{LoRA \cite{hu2022lora}} & \mctwol{} & \mctwo{\textbf{W} will be \textbf{plugged} into \textbf{SelfAttn layers}} \\
\multicolumn{2}{c|}{BitFit \cite{ben-zaken-etal-2022-bitfit}} & \mctwol{\multirow{1}{*}{}} & \mctwo{\textbf{W} will be \textbf{add} into \textbf{Bias terms}}\\
\bottomrule
\end{tabular}
}
\end{adjustbox}
\caption{We uniformly re-frame the transformations of \PET methods as modifications $\Delta \mathbf{h}$ of specific hidden states in the corresponding PLM layer ($f$) where $\mathbf{W}$ is introduced in computing $\Delta \mathbf{h}$, as suggested by \citet{he2022towards,hu2022sparse}. Each \PET method has $p$ tunable weights $\mathbf{W}$ in designed positions. Hence, we represent each \PET tunable module as $\theta=\{\mathbf{W}_{1}, \mathbf{W}_{2},...,\mathbf{W}_{p}\}$.}
\label{table:unified_view}
\end{center} 
\end{table*}

Although these \PET methods have distinct tunable modules, they can be unified into a similar form. \citet{he2022towards} formalize \PET methods as a unified framework to study the connections among \PET methods. \citet{yi2022} also conduct the same study and further indicate that the optimization of different \PET methods can be unified in a similar subspace. In this paper, we leverage these unified perspectives to explain the impact of model scaling on \PET in the final discussion (\cref{sec:discussion}).


\paragraph{The Power of Model Scaling}
\label{par:the_power_of_model_scale}
With the scaling of model size, PLMs emerge numerous capabilities, including reasoning ability \cite{wei2022_chain_of_thought,wei2022emergent}, and can achieve state-of-the-art results in various understanding and generation tasks \cite{galm2021,palm2022}. 

In the adaption perspective, some researchers find that performing some \PET methods \cite{lester-etal-2021-power,deltatuning,transferabilitysu} on large-scale models can almost achieve the full-parameter fine-tuning performance. In this paper, we further find that as the model scale increases, the performance differences among distinct \PET methods become smaller (\cref{sec:experiment_and_analysis}). Hence, we study the impact of model scaling on \PET methods (\cref{sec:ablations}) to fathom this phenomenon and explain it from the optimization perspective (\cref{sec:discussion}).



\section{Preliminary}
\label{label:preliminary}
In this section, we first introduce the Transformer framework (\cref{ssec:transformer_framework}) and the most representative \PET (\cref{ssec:parameter_efficient_tuning_methods}).

\subsection{Transformer Framework}
\label{ssec:transformer_framework}
The Transformer model \cite{NIPS2017_3f5ee243} is the mainstream architecture for most powerful PLMs. The model is stacked of $L$ blocks, each of which consists of a sequence of layers, including self-attention and feed-forward network. During the forward pass through each block, the input hidden state is applied with the sequence of layers. For simplicity, we formalize the transformation of each layer as

\begin{equation}
\label{eq:module_transformation}
    \mathbf{h}^{out}=f(\mathbf{h}^{in}).
\end{equation}
Under the layer as the operator $f$, the input hidden state $\mathbf{h}^{in}\in\mathbb{R}^{s \times d_{in}}$ is transformed into the output hidden state $\mathbf{h}^{out}\in\mathbb{R}^{s \times d_{out}}$, where $s$ is the input length and $d_{in}$, $d_{out}$ are dimensions.

\subsection{Parameter Efficient Tuning (PET)}
\label{ssec:parameter_efficient_tuning_methods}
Different \PET methods\footnote{More implementation details are left in \cref{appendix:pet}.} are equipped with diverse modules $\theta$ as shown in \figref{fig:the_power_of_scale}. These modules are composed of tunable parameters $\mathbf{W}$ that modify the original layers and the corresponding transformations in PLMs. To make comparisons, we follow the unified view \cite{he2022towards,hu2022sparse} to re-frame the transformations of all \PET methods as the modifications $\Delta \mathbf{h}$ of specific hidden states in the corresponding PLM's layers as follows:
\begin{equation}
\label{eq:module_transformation_pet}
    \mathbf{h}^{out} = f(\mathbf{h}^{in})+\Delta \mathbf{h}.
\end{equation}

In the training process, given a downstream task $\mathcal{D}=\{X,Y\}$, we only optimize all tunable parameters of the module $\theta$ for each \PET method to generate desired outputs $Y$ of a downstream task while freezing the rest of the parameters $\Phi$ in a PLM $\mathcal{M}$, as shown in \figref{fig:the_power_of_scale}\footnote{The manipulations, including addition, concatenation, and plugging, are discussed in  \cref{ssec:afp_tuning_methods}.}. Formally, the training objective is to minimize $\mathcal{L}$ as follows:
\begin{equation}
\label{eq:pet_training_objective}
    \min\nolimits_{\theta}\mathcal{L}(\mathcal{M}_{(\Phi,\theta)}(X), Y).
\end{equation}

\section{Main Experiments}
\label{sec:experiment_and_analysis}
To explore the impact of model scaling on these \PET methods, we first introduce the investigated tasks, PLMs, and settings of the existing representative \PET methods in the experiments (\cref{para:training_details}), and then report the main experimental results (\cref{ssec:main_experiments}).

\subsection{Experimental Settings}
\label{ssec:experimental_settings}
\paragraph{Investigated NLP Tasks}
\label{para:investigated_nlp_tasks}
We investigate $11$ tasks, which can be divided into $5$ categories: (1) \textit{Sentiment Analysis} (SA), including SST-2 \cite{socher-etal-2013-recursive}, IMDB \cite{maas-EtAl:2011:ACL-HLT2011}, and Rotten Tomatoes \cite{pang-lee-2005-seeing}; (2) \textit{Natural Language Inference} (NLI), including MNLI \cite{williams-etal-2018-broad}, QNLI \cite{wang2018glue}, and RTE \cite{bos-markert-2005-recognising}; (3) \textit{Paraphrase Identification} (PI), including MRPC \cite{dolan-brockett-2005-automatically} and QQP \cite{sharma2019natural}; (4) \textit{Question Answering} (QA), including NQ-Open \cite{lee-etal-2019-latent}; (5) \textit{Summarization} (SUM), including SAMSum \cite{gliwa-etal-2019-samsum} and Multi-News \cite{fabbri-etal-2019-multi}. More details are in \cref{appendix:task_and_dataset}.

\paragraph{Investigated PLMs}
\label{para:investigated_plms}
We will experiment on three series of PLM backbones: BERT \cite{devlin2018bert}, BLOOM \cite{2023bloom}, and T5 \cite{raffel2020exploring} representing encoder-based model, decoder-based model, and sequence-to-sequence based model, respectively. Since BERT has fixed-length output limitation, we only investigate SA, PI, and NLI categories of tasks on it. Differently, BLOOM and T5 models have no fixed-length output limitation; thus, we investigate all tasks on them.

\paragraph{Training Details of \PET Methods}
\label{para:training_details}
We select four representative \PET methods: Prompt \cite{lester-etal-2021-power}, BitFit \cite{ben-zaken-etal-2022-bitfit}, Adapter \cite{pmlr-v97-houlsby19a}, and LoRA \cite{hu2022lora}, for conducting analysis experiments. To ensure the consistency of the \PET methods' performance, we maintain the original design of each method, including the positions of tunable parameters and the number of trainable parameters, as reported in the respective original papers. Additionally, we train each \PET method on $11$ tasks using $3$ different random seeds and report their average performance. Further details regarding the training configurations can be found in \cref{appendix:pet}.

\begin{figure}[!t]
\centering
\includegraphics[width=0.49\textwidth]{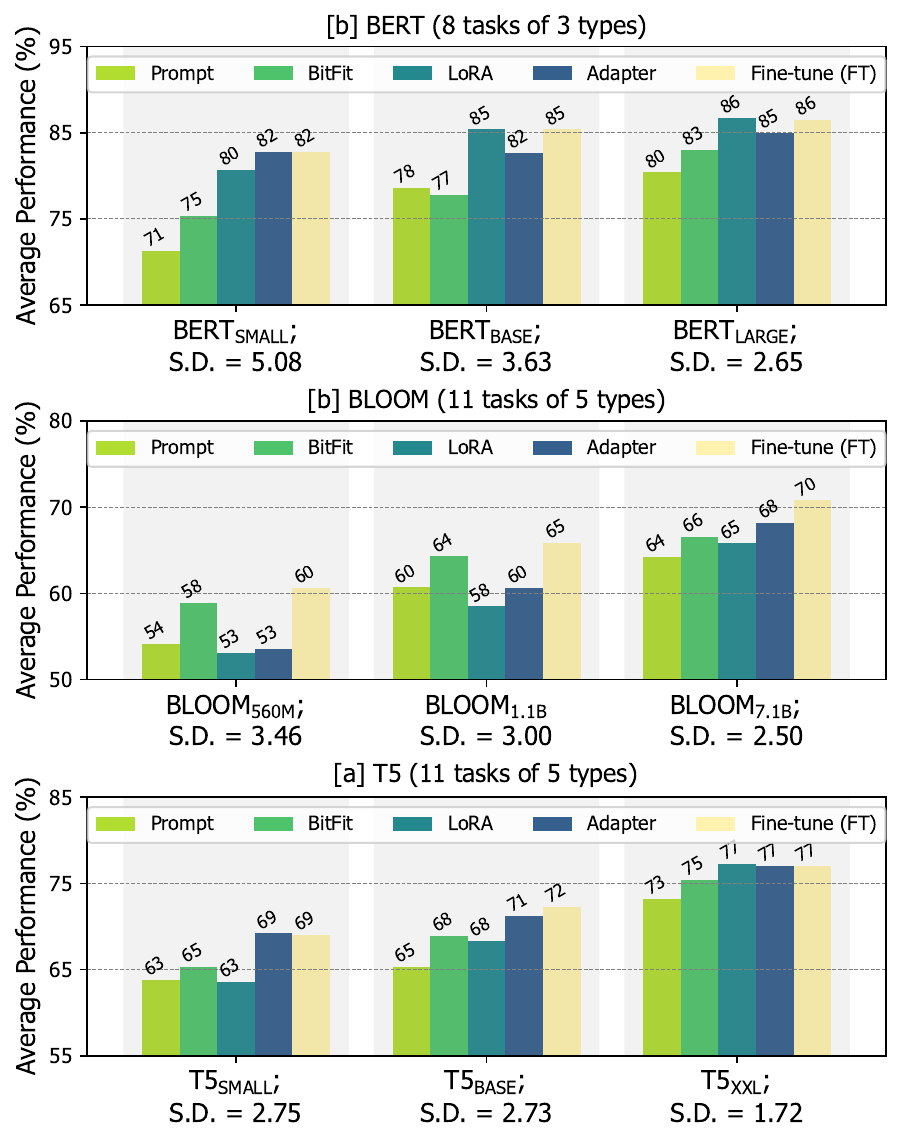}
\caption{We investigate the average performance of the tuning methods, including Prompt, BitFit, LoRA, Adapter, and full-parameter fine-tuning, on three series of models. As the model scaling increases, the performance differences (standard deviation (S.D.)) among tuning methods become smaller.} 
\label{fig:main_exp}
\end{figure}

\subsection{Model Scaling Impact on \PET Methods}
\label{ssec:main_experiments}

To investigate the impact of model scaling on \PET methods, we arrange the Pre-trained Language Models (PLMs) in ascending order based on their model scale, and we report the performance of \PET methods on each type of PLM.

Results are reported in \figref{fig:main_exp}. First, we can observe that the \PET methods exhibit noticeable performance differences (standard deviation (S.D.)) from each other on the general-scale models (\BERTSMALL and \BERTBASE in the sub-figure \texttt{[a]}; \BLOOMSMALL and \BLOOMBASE in the sub-figure \texttt{[b]}; \TSMALL and \TBASE in the sub-figure \texttt{[c]}). This phenomenon is intuitive and demonstrates the critical impact of design differences (the position and quantity of parameters in the tunable module) on the performance of \PET methods. This finding has been consistently found in numerous prior works \cite{deltatuning,hu2022sparse}.

However, we find that as the model scaling increases (from \BERTSMALL to \BERTLARGE in the sub-figure \texttt{[a]}; from \BLOOMSMALL to \BLOOMLARGE in the sub-figure \texttt{[b]}; from \TSMALL to \TXXL in the sub-figure \texttt{[c]}), the performance discrepancies among \PET methods diminish across all types of models, as evidenced by the decreasing standard deviation (S.D.) (from $5.08$ to $2.65$ on \texttt{[a]} BERT; from $3.46$ to $2.50$ on \texttt{[b]} BLOOM; from $2.75$ to $1.72$ on \texttt{[c]} T5). This finding implies that \textit{the larger model scaling can mitigate the impact of the design differences among the \PET methods on performance}.



\section{Ablation Analyses}
\label{sec:ablations}

The design differences among the PET methods mainly lie in the tunable module's parameter position and parameter quantity. To further verify whether the model scaling will respectively remove the effects of the above differences on \PET methods, we conducted two ablations to investigate whether model scaling can mitigate (1) the impact of tunable parameter position and (2) the impact of tunable parameter quantity. 

However, only investigating the above four respective \PET methods is insufficient to cover enough variations of parameter position for
ablation study (1). This limitation makes us hard to preciously control the number of tunable parameters at the fine-grained (parameter level) in ablation study (2). Hence, we develop a more flexible \PET method, \textbf{A}rbitrary \textbf{P}arameter-\textbf{E}fficient \textbf{T}uning (\APET) method. Its tunable module can be arbitrary structure (\cref{ssec:afp_tuning_methods}) that facilitates us to explore various parameter positions in the ablation study (\cref{ssec:ms_model_structure}) and easier control the number of tunable parameters in the ablation study (\cref{ssec:ms_trainable_parameter}).

\subsection{Arbitrarily Parameter-Efficient Tuning (\APET)}
\label{ssec:afp_tuning_methods}

Similar to \PET methods, the \APET method is equipped with arbitrary module $\theta$ which is composed of $L$ tunable weights $\mathbf{W}$ distributed in any position of a model. Here, \APET have three operations to insert the tunable weight $\mathbf{W}$ into any position of the PLM, thereby modify the specific layers and their corresponding transformations as follows:

\paragraph{ADD}
\label{}
The tunable weight $\mathbf{W}$ will be into the PLM layer. The corresponding transformation of a PLM layer can be denoted as:
\begin{equation}
\label{eq:afp_add}
    \mathbf{h}^{out} = f(\mathbf{h}^{in}) + \mathbf{W}_{\texttt{1}}.
\end{equation}

\paragraph{CONCAT}
\label{}
The tunable weight $\mathbf{W}$ will be concatenated with the hidden state or the layer in the PLM. The corresponding transformation of a PLM layer can be denoted as:
\begin{equation}
\label{eq:afp_concatenate}
\begin{aligned}
     \mathbf{h}^{out} = f(\mathbf{h}^{in}) +
     \left\{\begin{matrix}
     f(\mathbf{W}_{\texttt{2}}) \\ 
     \alpha \mathbf{h}^{in} \mathbf{W}_{\texttt{3}} \mathbf{W}_{\texttt{4}}
    \end{matrix}\right.
\end{aligned}
\end{equation}

\begin{figure}[!t]
\centering
\includegraphics[width=0.485\textwidth]{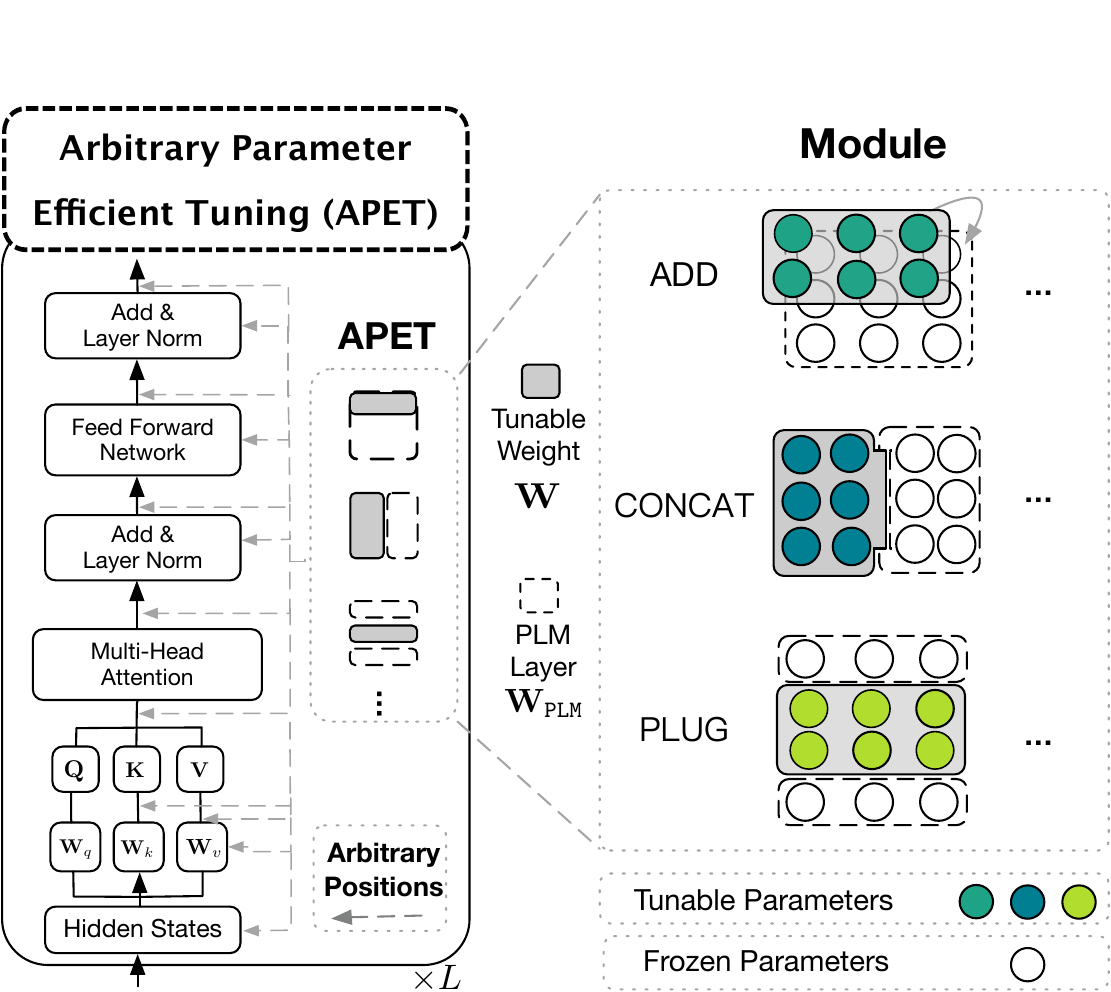}
\caption{The tunable modules of \APET methods ($\theta= \{ \mathbf{W}_{1}, \mathbf{W}_{2}, ..., \mathbf{W}_{L} \}$) are composed $L$ tunable weights $\mathbf{W}$ with arbitrary structures. There are three operations (ADD, CONCAT, PLUG) for inserting these tunable weights $\mathbf{W}$ into a PLM.} 
\label{fig:distribute_structure}
\end{figure}

\paragraph{PLUG}
\label{}
The tunable weight $\mathbf{W}$ will be plugged between PLM layers. The corresponding transformation of a PLM layer can be denoted as:
\begin{equation}
\label{eq:afp_plugin}
    \mathbf{h}^{out} = f(\mathbf{h}^{in}) + \sigma(f(\mathbf{h}^{in}) \mathbf{W}_{\texttt{5}}) \mathbf{W}_{\texttt{6}}.
\end{equation}

Note that the inserted tunable weights $\mathbf{W}$ are not limited to the aforementioned structure as shown in \figref{fig:distribute_structure}; they can be arbitrary structures. According to the inserted tunable weights and the corresponding modifications, the transformations of a PLM layer for \APET method can be expressed as:
\begin{equation}
\label{eq:afp_complete}
\begin{aligned}
     \mathbf{h}^{out} =& f(\mathbf{h}^{in}) + 
     \left\{\begin{matrix}
     \mathbf{W}_{\texttt{1}} \\
     f(\mathbf{W}_{\texttt{2}}) \\ 
     \alpha \mathbf{h}^{in} \mathbf{W}_{\texttt{3}} \mathbf{W}_{\texttt{4}} \\
     \sigma(f(\mathbf{h}^{in}) \mathbf{W}_{\texttt{5}}) \mathbf{W}_{\texttt{6}} \\
     \vdots
    \end{matrix}\right. 
\end{aligned}
\end{equation}
By comparing Equation \ref{eq:afp_complete} with the equations of previously introduced Equation \ref{eq:module_transformation_pet}, it is obvious that the \PET methods are special cases of \APET method. 

\begin{figure*}[!t]
\centering
\includegraphics[width=1\textwidth]{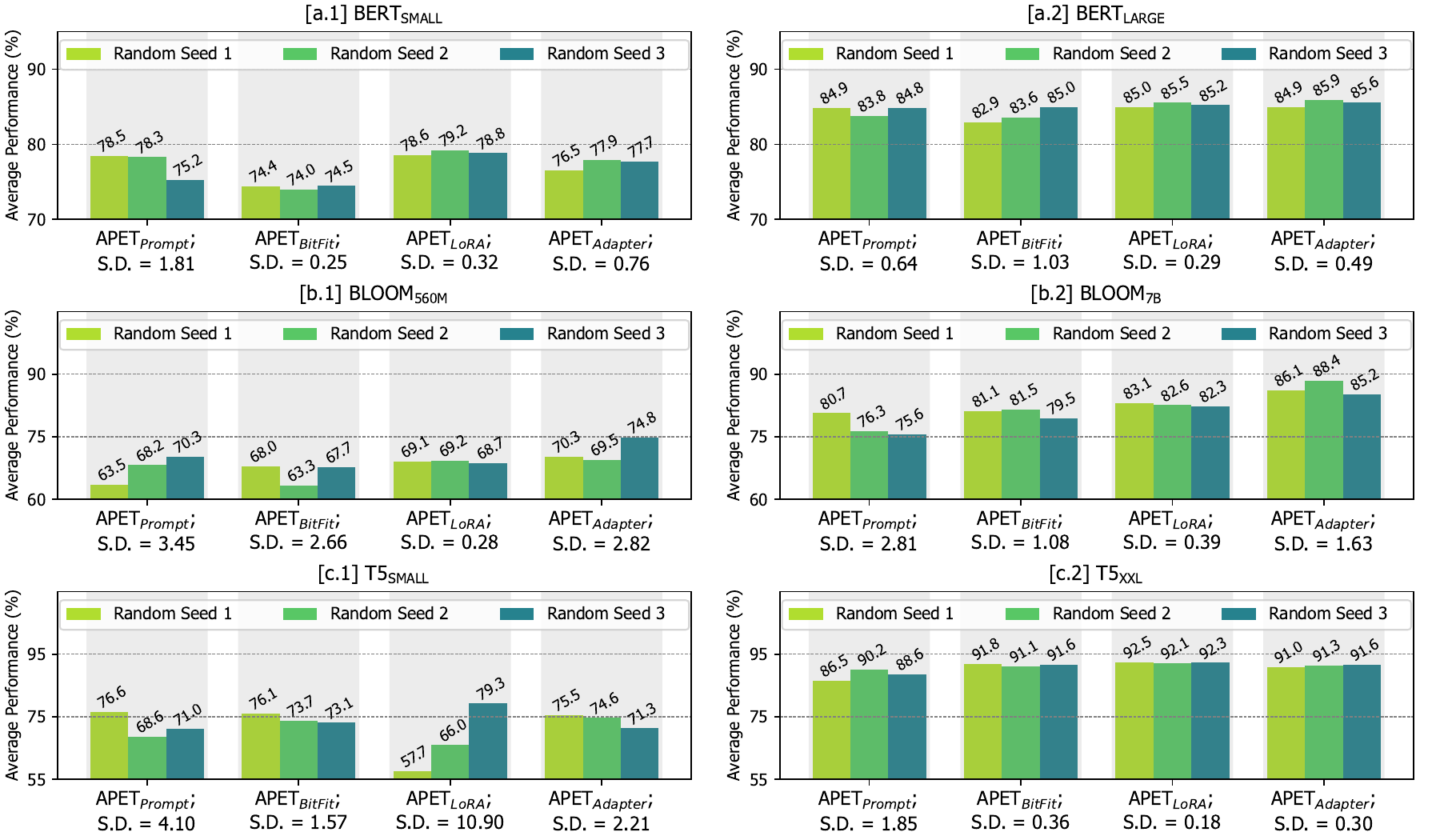}
\caption{The parameter quantity in each group (bar: \colorbox{lightgray}{\textcolor{lightgray}{h}}) corresponds to the aforementioned four \PET methods'. We denote the \APET methods with the coresponding numbers of parameters as $\text{APET}_{\text{Prompt}}$, $\text{APET}_{\text{BitFit}}$, $\text{APET}_{\text{LoRA}}$, and $\text{APET}_{\text{Adapter}}$, respectively. Each \APET method will arbitrarily select tunable parameters with different random seeds, each random seed representing a different parameter distribution. Here, S.D. means the standard deviation. As the model scaling increases, the impact caused by the parameter position on the performance becomes minor.} 
\label{fig:structure_and_position}
\end{figure*}


The module $\theta$ of \APET are composed of arbitrarily inserted weights $\mathbf{W}$, which can be expressed as $\theta= \{\mathbf{\widehat{W}}_{1}, \mathbf{\widehat{W}}_{2}, ..., \mathbf{\widehat{W}}_{L} \}$. In the training process, we follow Equation \eqref{eq:pet_training_objective} only to optimize $\theta$ while freezing the rest of the parameters ($\Phi$) in a PLM.

\subsection{The Impact of Differences in Parameter Position on Performance}
\label{ssec:ms_model_structure}



To investigate whether model scaling can mitigate the impact of parameter position in \PET, we initially freeze other significant factors, i.e., the number of tunable parameters, that could potentially affect the performance. Given that the tunable parameters of the four aforementioned \PET methods are fixed in the same positions, it is challenging for us to precisely conduct an experiment to assess the impact of position. Under this limitation, we then employ the \APET method to arbitrarily select tunable parameters with different random seeds, each random seed representing a different parameter distribution, and train them on the tasks.

In the experiments, we set the number of tunable parameters for the \APET methods in four groups. The parameter quantity in each group (bar: \colorbox{lightgray}{\textcolor{lightgray}{h}}) corresponds to that of the aforementioned four \PET methods' (Prompt, BitFit, LoRA, Adapter). We denote these \APET methods with varying numbers of parameters\footnote{The number of tunable parameters are left in \cref{appendix:apet_parameter_number}.} as $\text{APET}_{\text{Prompt}}$, $\text{APET}_{\text{BitFit}}$, $\text{APET}_{\text{LoRA}}$, and $\text{APET}_{\text{Adapter}}$, respectively. Besides, we conduct the ablation study on three series of models (BERT, BLOOM, and T5) and report task (SST, RTE, and MRPC) average performance.



\paragraph{Performance Comparison}
\label{}
As shown in \figref{fig:structure_and_position}, there are four groups of comparisons in each sub-graph. We can observe that as a PLM size scales (BERT: from \texttt{[a.1]} to \texttt{[a.2]}; BLOOM: from \texttt{[b.1]} to \texttt{[b.2]}; T5: from \texttt{[c.1]} to \texttt{[c.2]}), the performance differences (standard deviation (S.D)) of \APET methods \textbf{within each group} decrease. Based on this findings, we argue that \textit{larger models demonstrate greater effectiveness in mitigating the impact of differences in parameter position on performance}.

In addition, we have observed that despite the different number of tunable parameters \textbf{in four different groups} (bar: \colorbox{lightgray}{\textcolor{lightgray}{h}}) of \APET methods, they have fewer performance differences on the larger model. We will delve into this finding further and provide an explanation for this phenomenon in \cref{ssec:ms_trainable_parameter}.




\begin{figure*}[!th]
\centering
\includegraphics[width=1\textwidth]{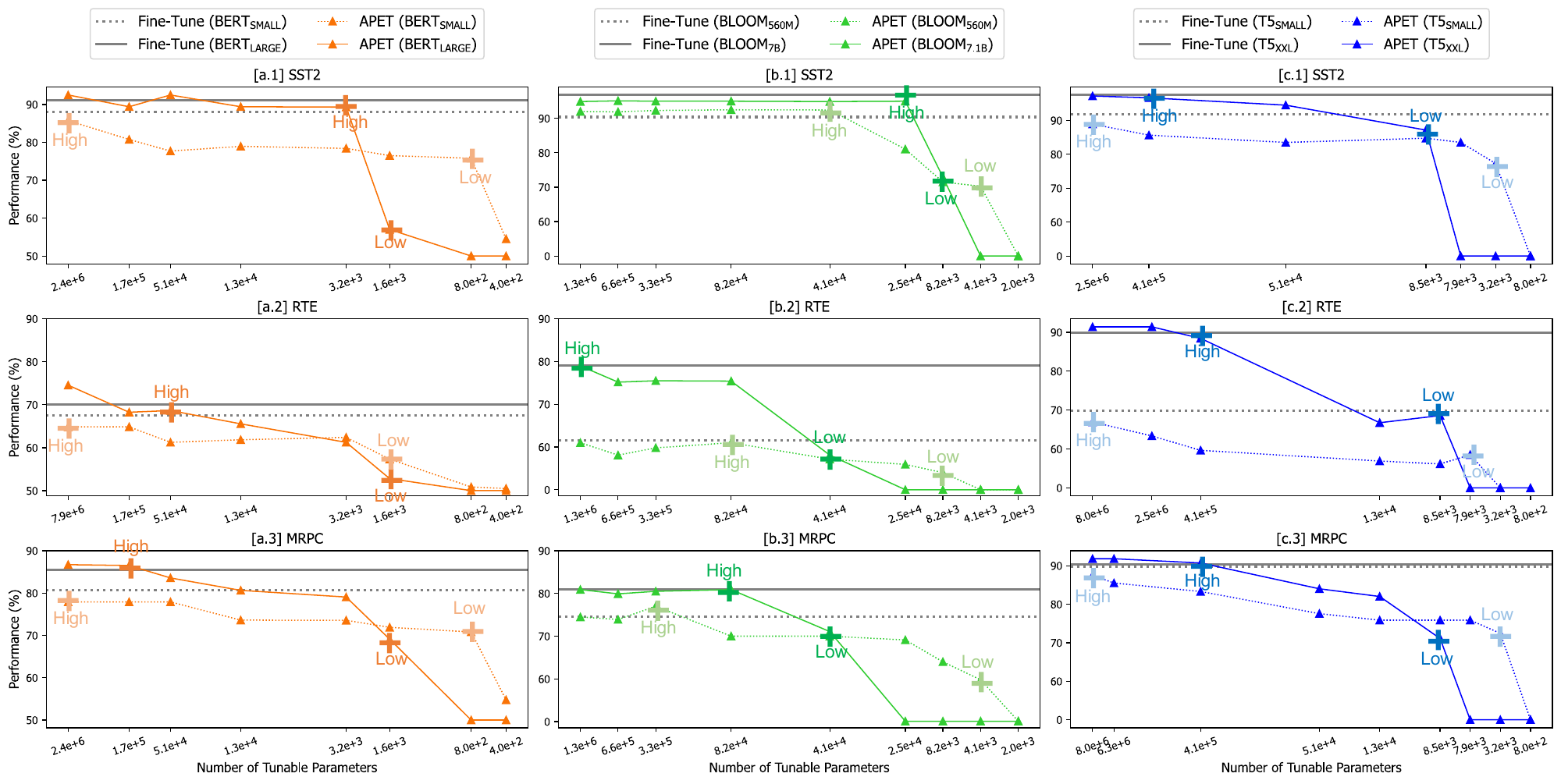}
\caption{Given the different numbers of tunable parameters, we observe \APET performance on three series of models and tasks. We find that (1) the model scaling can make tuning methods optimize fewer necessarily tuned parameters to reach full-parameter fine-tuning performance (\textcolor{gray}{\textbf{- - -}} and \textcolor{gray}{\textbf{-----}}); (2) \APET methods require the similiar number of tunable parameters (low parameter thresholds lie the similiar range) to exceed random guess performance on the same models.}
\label{fig:ratio_threshold}
\end{figure*}

\subsection{The Impact of Differences in The Number of Tunable Parameters on Performance}
\label{ssec:ms_trainable_parameter}

In this section, given the \APET method under different numbers of tunable parameters, we observe their performance to conduct an ablation study.

From the reported results in \figref{fig:ratio_threshold}, we can find that (1) on the smaller models, e.g., \BERTSMALL (\textcolor{def_deep_orange}{\textbf{- - -}}), \BLOOMSMALL (\textcolor{gr}{\textbf{- - -}}), \TSMALL (\textcolor{def_deep_blue}{\textbf{- - -}}) when the tunable parameters of tuning methods are fewer than a certain number, the performance will drop to randomly guess performance; (2) similarly, this phenomenon still holds on the larger models, \BERTLARGE (\textcolor{def_deep_orange}{\textbf{-----}}), \BLOOMLARGE (\textcolor{gr}{\textbf{-----}}), \TXXL (\textcolor{def_deep_blue}{\textbf{-----}}). Based on these findings, we can argue that that \textit{model scaling cannot adequately eliminate the impact of the number of tunable parameters on the performance of \PET methods}.

Interestingly, we find two parameter thresholds for tunable parameters in all models and name them as low parameter threshold (\textcolor{def_deep_orange}{Low}, \textcolor{def_deep_orange_light}{Low}, \textcolor{gr}{Low}, \textcolor{gr_light}{Low}, \textcolor{def_deep_blue}{Low}, \textcolor{def_deep_blue_light}{Low}) for necessary tuned parameters and the high parameter threshold (\textcolor{def_deep_orange}{High}, \textcolor{def_deep_orange_light}{High}, \textcolor{gr}{High}, \textcolor{gr_light}{High}, \textcolor{def_deep_blue}{High}, \textcolor{def_deep_blue_light}{High}) for necessary tuned parameters, respectively in Figure \ref{fig:ratio_threshold}. When tunable parameters are more than \textit{low parameter threshold}, the \APET method can \textit{exceed random performance} (e.g., $\frac{1\times 100}{\text{Number of label types}}\%$ on BERT, 0\% on BLOOM, and 0\% on T5); when the tunable parameters are more than high parameter threshold, the \APET method can almost achieve the full-parameter fine-tuning (FT) performance. Furthermore, we find that the model scaling affects the two parameter thresholds. Hence, we explore this phenomenon in the following paragraphs.




\paragraph{High Threshold of Necessary Tuned Parameters}
\label{}


Based on the experimental results in the sub-graph \texttt{[c.1]} (SST2) of \figref{fig:ratio_threshold}, we find that the high threshold of the larger model is consistently lower than the high threshold of the smaller model. This phenomenon holds true across all tasks (\texttt{SST2}, \texttt{RTE}, \texttt{MRPC}), and for all series of models, as depicted in all sub-graphs. Therefore, we can conclude that \textit{model scaling enables tuning methods to train fewer necessary parameters while achieving the similar performance of full-parameter fine-tuning}.


This conclusion can intuitively explain why \APET methods can achieve relatively similar performance on larger models, especially on \TXXL, as illustrated in the aforementioned [c.2] in \figref{fig:main_exp}. This is due to the fact that the number of tunable parameters in each group of \APET methods surpasses the high parameter thresholds on \TXXL; hence, they all achieve the similar performance of full-parameter fine-tuning.


\paragraph{Low Threshold of Necessary Tuned Parameters}
\label{}
From the above results, we find that \APET methods will exceed the random guess performance (0\% on T5; 0\% on BLOOM; 50\% on BERT) and immediately reach the 80\textasciitilde90\% full-parameter fine-tuning performance when the tunable parameters are more than low thresholds. However, the low thresholds are relatively higher on larger models (\BERTLARGE, \BLOOMLARGE, \TXXL). Namely, \APET methods require more tunable parameters to exceed the random guess performance. This phenomenon is consistent over all tasks on all series of models. Hence, we can infer that the model scaling cannot reduce the number of necessary tuned parameters to drive PLMs to perform downstream tasks.

Furthermore, it is worth noting that \textit{the low parameter thresholds of the \APET methods almost lie in the same range on the same models}. Specifically, the range of low thresholds are in [\texttt{8.0e+2}, \texttt{3.2e+3}] on \BERTLARGE, and [\texttt{4.0e+2}, \texttt{1.6e+3}] on \BERTSMALL; [\texttt{8.2e+3}, \texttt{4.1e+4}] on \BLOOMLARGE, [\texttt{8.2e+3}, \texttt{4.1e+3}] on \BLOOMSMALL; [\texttt{7.9e+3}, \texttt{8.5e+3}] on \TXXL, [\texttt{8.0e+2}, \texttt{7.9e+3}] on \TSMALL. We will explain this phenomenon from the optimization perspective in \cref{sec:discussion}.

\section{Discussing the Ablation Results from the Optimization Perspectives}
\label{sec:discussion}


The objectives of all parameter-efficient tuning methods (\PET, \APET) can be expressed as $\min\nolimits_{\theta}\mathcal{L}(\mathcal{M}_{(\Phi,\theta)}(X), Y)$ as introduced in Equation \eqref{eq:pet_training_objective}, where $\theta$ is a tunable module. The module $\theta$ of different \PET methods consists of different structures and varying numbers of tunable parameters. In this paper, we investigate the impact of model scaling on different modules, which possess varying numbers of tunable parameters distributed across multiple positions. We find that the larger model scaling can (1) mitigate the effects caused by the difference positions of tunable parameters (\cref{ssec:ms_model_structure}) and (2) make \PET methods optimize fewer tunable parameters to achieve full-parameter fine-tuning performance (\cref{ssec:ms_trainable_parameter}). To further fathom these phenomena, we will investigate the underlying reasons from an optimization perspective. (3) Besides, we also observe that \PET methods can optimize almost the similar number of necessarily tuned parameters to exceed random guess performance on the same backbone models (\cref{ssec:ms_trainable_parameter}). Although phenomenon (3) is not caused by model scaling, we can also explain it from the optimization perspective. Next, we together discuss it and the above two findings (1) and (2) in the following paragraphs.

\paragraph{Why model scaling mitigates the effects caused by the differences in positions of tunable parameters on the \PET performance?}
\label{}
From the optimal control perspective, a tunable module ($\theta$) of a tuning method can be seen as a controller \cite{yang2022on,deltatuning} to drive PLMs towards downstream tasks. As the model scale increases, the larger model has higher parameter redundancy \cite{aghajanyan-etal-2021-intrinsic}, allowing arbitrary selection of tunable parameters for tuning without greatly degrading performance \cite{desai-etal-2019-evaluating,NEURIPS2020_b6af2c97,prasanna-etal-2020-bert,pmlr-v119-evci20a}; thus, controllers (modules) might have higher degrees of freedom. 

This might explain why the aribitray positions of the tunable parameters have less impact such that all \PET methods can achieve the similar performance on the larger models. It is worth noting that even though the distribution of tunable parameters have less impact on the performance, it still affects converge speeds. Thus, finding a better parameter distribution to improve the converge speeds for \PET methods is a direction worthy of exploring.


\paragraph{Why model scaling leverages the fewer tunable parameters to achieve full-parameter fine-tuning performance?}
\label{}
Tuning $\theta$ to steer a PLM towards downstream NLP tasks can be seen as adaptations. From the perspective of representation space, the adaptations of \PET methods can be re-parameterized into a unified low dimensional subspace \cite{qin2021exploring, aghajanyan-etal-2021-intrinsic,yi2022}. \citet{aghajanyan-etal-2021-intrinsic} further demonstrate that adaptation on a larger PLM can be re-parameterized into the lower dimensional space; this implicitly explains why \PET methods can optimize fewer parameters on larger-scale models, e.g., \TXXL, to meet the full-parameter fine-tuning performance on tasks.

\paragraph{Why can \PET methods optimize the similar numbers of tunable parameters to exceed random guessing?}
\label{}
As stated above, the adaptations of the \PET methods can be re-parameterized into a unified subspace. \citet{qin2021exploring} show that this low dimensional subspace is shared among all NLP tasks for the same \PET methods. \citet{yi2022} further suggest that this subspace is also shared among various \PET methods. This might implicitly explain why all \PET methods can tune the similar numbers of necessary tuned parameters to exceed the random guessing performance on the same models, even for the different tasks (\cref{ssec:ms_trainable_parameter}).

\section{Conclusion}
\label{sec:conclusion}

The realm of model scaling for LLMs presents important and intriguing directions for the LLM community. The increasing of model scale unveils numerous emerging capabilities and advantages. In this work, our primary emphasis is on the impact of model scaling as it pertains to PET methods. Through our comprehensive observation studies and in-depth discussions from optimization perspectives, we gain deeper insights into the effects of model scaling on PET and the reasons behind the observed phenomena. We believe that our findings will serve as a catalyst, inspiring further meticulous research and exploration in this area.


\section{Limitations}
\label{sec:limitations}

This paper might have some possible limitations as follows: (1) we only explore the effects of the scaling law on performance. There might be other research points worth exploring, such as the power of model scale to convergence speed; (2) we study the power of model scale with comprehensive empirical experiments and explain the findings from the optimization perspective. There might be more theoretical proofs to explain these exciting findings.




\bibliography{custom}

\begin{thebibliography}{49}
\expandafter\ifx\csname natexlab\endcsname\relax\def\natexlab#1{#1}\fi

\bibitem[{Aghajanyan et~al.(2021)Aghajanyan, Gupta, and
  Zettlemoyer}]{aghajanyan-etal-2021-intrinsic}
Armen Aghajanyan, Sonal Gupta, and Luke Zettlemoyer. 2021.
\newblock \href {https://doi.org/10.18653/v1/2021.acl-long.568} {Intrinsic
  dimensionality explains the effectiveness of language model fine-tuning}.
\newblock In \emph{Proceedings of the 59th Annual Meeting of the Association
  for Computational Linguistics and the 11th International Joint Conference on
  Natural Language Processing (Volume 1: Long Papers)}, pages 7319--7328,
  Online. Association for Computational Linguistics.

\bibitem[{Ben~Zaken et~al.(2022)Ben~Zaken, Goldberg, and
  Ravfogel}]{ben-zaken-etal-2022-bitfit}
Elad Ben~Zaken, Yoav Goldberg, and Shauli Ravfogel. 2022.
\newblock \href {https://doi.org/10.18653/v1/2022.acl-short.1} {{B}it{F}it:
  Simple parameter-efficient fine-tuning for transformer-based masked
  language-models}.
\newblock In \emph{Proceedings of the 60th Annual Meeting of the Association
  for Computational Linguistics (Volume 2: Short Papers)}, pages 1--9, Dublin,
  Ireland. Association for Computational Linguistics.

\bibitem[{Bos and Markert(2005)}]{bos-markert-2005-recognising}
Johan Bos and Katja Markert. 2005.
\newblock \href {https://aclanthology.org/H05-1079} {Recognising textual
  entailment with logical inference}.
\newblock In \emph{Proceedings of Human Language Technology Conference and
  Conference on Empirical Methods in Natural Language Processing}, pages
  628--635, Vancouver, British Columbia, Canada. Association for Computational
  Linguistics.

\bibitem[{Chen et~al.(2023)Chen, Zhang, Shi, Li, Smola, and
  Yang}]{chen2023parameterefficient}
Jiaao Chen, Aston Zhang, Xingjian Shi, Mu~Li, Alex Smola, and Diyi Yang. 2023.
\newblock \href {https://openreview.net/forum?id=XSRSWxyJIC}
  {Parameter-efficient fine-tuning design spaces}.
\newblock In \emph{The Eleventh International Conference on Learning
  Representations}.

\bibitem[{Chen et~al.(2020)Chen, Frankle, Chang, Liu, Zhang, Wang, and
  Carbin}]{NEURIPS2020_b6af2c97}
Tianlong Chen, Jonathan Frankle, Shiyu Chang, Sijia Liu, Yang Zhang, Zhangyang
  Wang, and Michael Carbin. 2020.
\newblock \href
  {https://proceedings.neurips.cc/paper/2020/hash/b6af2c9703f203a2794be03d443af2e3-Abstract.html}
  {The lottery ticket hypothesis for pre-trained {BERT} networks}.
\newblock In \emph{Advances in Neural Information Processing Systems 33: Annual
  Conference on Neural Information Processing Systems 2020, NeurIPS 2020,
  December 6-12, 2020, virtual}.

\bibitem[{Chowdhery et~al.(2022)Chowdhery, Narang, Devlin, Bosma, Mishra,
  Roberts, Barham, Chung, Sutton, Gehrmann, Schuh, Shi, Tsvyashchenko, Maynez,
  Rao, Barnes, Tay, Shazeer, Prabhakaran, Reif, Du, Hutchinson, Pope, Bradbury,
  Austin, Isard, Gur-Ari, Yin, Duke, Levskaya, Ghemawat, Dev, Michalewski,
  Garcia, Misra, Robinson, Fedus, Zhou, Ippolito, Luan, Lim, Zoph, Spiridonov,
  Sepassi, Dohan, Agrawal, Omernick, Dai, Pillai, Pellat, Lewkowycz, Moreira,
  Child, Polozov, Lee, Zhou, Wang, Saeta, Diaz, Firat, Catasta, Wei,
  Meier-Hellstern, Eck, Dean, Petrov, and Fiedel}]{palm2022}
Aakanksha Chowdhery, Sharan Narang, Jacob Devlin, Maarten Bosma, Gaurav Mishra,
  Adam Roberts, Paul Barham, Hyung~Won Chung, Charles Sutton, Sebastian
  Gehrmann, Parker Schuh, Kensen Shi, Sasha Tsvyashchenko, Joshua Maynez,
  Abhishek Rao, Parker Barnes, Yi~Tay, Noam Shazeer, Vinodkumar Prabhakaran,
  Emily Reif, Nan Du, Ben Hutchinson, Reiner Pope, James Bradbury, Jacob
  Austin, Michael Isard, Guy Gur-Ari, Pengcheng Yin, Toju Duke, Anselm
  Levskaya, Sanjay Ghemawat, Sunipa Dev, Henryk Michalewski, Xavier Garcia,
  Vedant Misra, Kevin Robinson, Liam Fedus, Denny Zhou, Daphne Ippolito, David
  Luan, Hyeontaek Lim, Barret Zoph, Alexander Spiridonov, Ryan Sepassi, David
  Dohan, Shivani Agrawal, Mark Omernick, Andrew~M. Dai,
  Thanumalayan~Sankaranarayana Pillai, Marie Pellat, Aitor Lewkowycz, Erica
  Moreira, Rewon Child, Oleksandr Polozov, Katherine Lee, Zongwei Zhou, Xuezhi
  Wang, Brennan Saeta, Mark Diaz, Orhan Firat, Michele Catasta, Jason Wei,
  Kathy Meier-Hellstern, Douglas Eck, Jeff Dean, Slav Petrov, and Noah Fiedel.
  2022.
\newblock \href {https://arxiv.org/abs/2204.02311} {Palm: Scaling language
  modeling with pathways}.

\bibitem[{Desai et~al.(2019)Desai, Zhan, and Aly}]{desai-etal-2019-evaluating}
Shrey Desai, Hongyuan Zhan, and Ahmed Aly. 2019.
\newblock \href {https://doi.org/10.18653/v1/D19-6117} {Evaluating lottery
  tickets under distributional shifts}.
\newblock In \emph{Proceedings of the 2nd Workshop on Deep Learning Approaches
  for Low-Resource NLP (DeepLo 2019)}, pages 153--162, Hong Kong, China.
  Association for Computational Linguistics.

\bibitem[{Devlin et~al.(2019)Devlin, Chang, Lee, and
  Toutanova}]{devlin2018bert}
Jacob Devlin, Ming-Wei Chang, Kenton Lee, and Kristina Toutanova. 2019.
\newblock \href {https://doi.org/10.18653/v1/N19-1423} {{BERT}: Pre-training of
  deep bidirectional transformers for language understanding}.
\newblock In \emph{Proceedings of the 2019 Conference of the North {A}merican
  Chapter of the Association for Computational Linguistics: Human Language
  Technologies, Volume 1 (Long and Short Papers)}, pages 4171--4186,
  Minneapolis, Minnesota. Association for Computational Linguistics.

\bibitem[{Ding et~al.(2023)Ding, Qin, Yang, Wei, Yang, Su, Hu, Chen, Chan,
  Chen, Yi, Zhao, Wang, Liu, Zheng, Chen, Liu, Tang, Li, and Sun}]{deltatuning}
Ning Ding, Yujia Qin, Guang Yang, Fuchao Wei, Zonghan Yang, Yusheng Su,
  Shengding Hu, Yulin Chen, Chi-Min Chan, Weize Chen, Jing Yi, Weilin Zhao,
  Xiaozhi Wang, Zhiyuan Liu, Hai-Tao Zheng, Jianfei Chen, Yang Liu, Jie Tang,
  Juanzi Li, and Maosong Sun. 2023.
\newblock \href {https://www.nature.com/articles/s42256-023-00626-4#citeas}
  {Parameter-efficient fine-tuning of large-scale pre-trained language models}.
\newblock \emph{Nature Machine Intelligence}.

\bibitem[{Dolan and Brockett(2005)}]{dolan-brockett-2005-automatically}
William~B. Dolan and Chris Brockett. 2005.
\newblock \href {https://aclanthology.org/I05-5002} {Automatically constructing
  a corpus of sentential paraphrases}.
\newblock In \emph{Proceedings of the Third International Workshop on
  Paraphrasing ({IWP}2005)}.

\bibitem[{Du et~al.(2022)Du, Huang, Dai, Tong, Lepikhin, Xu, Krikun, Zhou, Yu,
  Firat, Zoph, Fedus, Bosma, Zhou, Wang, Wang, Webster, Pellat, Robinson,
  Meier{-}Hellstern, Duke, Dixon, Zhang, Le, Wu, Chen, and Cui}]{galm2021}
Nan Du, Yanping Huang, Andrew~M. Dai, Simon Tong, Dmitry Lepikhin, Yuanzhong
  Xu, Maxim Krikun, Yanqi Zhou, Adams~Wei Yu, Orhan Firat, Barret Zoph, Liam
  Fedus, Maarten~P. Bosma, Zongwei Zhou, Tao Wang, Yu~Emma Wang, Kellie
  Webster, Marie Pellat, Kevin Robinson, Kathleen~S. Meier{-}Hellstern, Toju
  Duke, Lucas Dixon, Kun Zhang, Quoc~V. Le, Yonghui Wu, Zhifeng Chen, and
  Claire Cui. 2022.
\newblock \href {https://proceedings.mlr.press/v162/du22c.html} {Glam:
  Efficient scaling of language models with mixture-of-experts}.
\newblock In \emph{International Conference on Machine Learning, {ICML} 2022,
  17-23 July 2022, Baltimore, Maryland, {USA}}, volume 162 of \emph{Proceedings
  of Machine Learning Research}, pages 5547--5569. {PMLR}.

\bibitem[{Evci et~al.(2020)Evci, Gale, Menick, Castro, and
  Elsen}]{pmlr-v119-evci20a}
Utku Evci, Trevor Gale, Jacob Menick, Pablo~Samuel Castro, and Erich Elsen.
  2020.
\newblock \href {http://proceedings.mlr.press/v119/evci20a.html} {Rigging the
  lottery: Making all tickets winners}.
\newblock In \emph{Proceedings of the 37th International Conference on Machine
  Learning, {ICML} 2020, 13-18 July 2020, Virtual Event}, volume 119 of
  \emph{Proceedings of Machine Learning Research}, pages 2943--2952. {PMLR}.

\bibitem[{Fabbri et~al.(2019)Fabbri, Li, She, Li, and
  Radev}]{fabbri-etal-2019-multi}
Alexander Fabbri, Irene Li, Tianwei She, Suyi Li, and Dragomir Radev. 2019.
\newblock \href {https://doi.org/10.18653/v1/P19-1102} {Multi-news: A
  large-scale multi-document summarization dataset and abstractive hierarchical
  model}.
\newblock In \emph{Proceedings of the 57th Annual Meeting of the Association
  for Computational Linguistics}, pages 1074--1084, Florence, Italy.
  Association for Computational Linguistics.

\bibitem[{Gliwa et~al.(2019)Gliwa, Mochol, Biesek, and
  Wawer}]{gliwa-etal-2019-samsum}
Bogdan Gliwa, Iwona Mochol, Maciej Biesek, and Aleksander Wawer. 2019.
\newblock \href {https://doi.org/10.18653/v1/D19-5409} {{SAMS}um corpus: A
  human-annotated dialogue dataset for abstractive summarization}.
\newblock In \emph{Proceedings of the 2nd Workshop on New Frontiers in
  Summarization}, pages 70--79, Hong Kong, China. Association for Computational
  Linguistics.

\bibitem[{Guo et~al.(2021)Guo, Rush, and Kim}]{guo2020parameter}
Demi Guo, Alexander Rush, and Yoon Kim. 2021.
\newblock \href {https://doi.org/10.18653/v1/2021.acl-long.378}
  {Parameter-efficient transfer learning with diff pruning}.
\newblock In \emph{Proceedings of the 59th Annual Meeting of the Association
  for Computational Linguistics and the 11th International Joint Conference on
  Natural Language Processing (Volume 1: Long Papers)}, pages 4884--4896,
  Online. Association for Computational Linguistics.

\bibitem[{He et~al.(2022{\natexlab{a}})He, Zhou, Ma, Berg{-}Kirkpatrick, and
  Neubig}]{he2022towards}
Junxian He, Chunting Zhou, Xuezhe Ma, Taylor Berg{-}Kirkpatrick, and Graham
  Neubig. 2022{\natexlab{a}}.
\newblock \href {https://openreview.net/forum?id=0RDcd5Axok} {Towards a unified
  view of parameter-efficient transfer learning}.
\newblock In \emph{The Tenth International Conference on Learning
  Representations, {ICLR} 2022, Virtual Event, April 25-29, 2022}.
  OpenReview.net.

\bibitem[{He et~al.(2022{\natexlab{b}})He, Ding, Dong, Zhang, and
  Tao}]{sparseadapter2022}
Shwai He, Liang Ding, Daize Dong, Jeremy Zhang, and Dacheng Tao.
  2022{\natexlab{b}}.
\newblock \href {https://aclanthology.org/2022.findings-emnlp.160}
  {{S}parse{A}dapter: An easy approach for improving the parameter-efficiency
  of adapters}.
\newblock In \emph{Findings of the Association for Computational Linguistics:
  EMNLP 2022}, pages 2184--2190, Abu Dhabi, United Arab Emirates. Association
  for Computational Linguistics.

\bibitem[{Houlsby et~al.(2019{\natexlab{a}})Houlsby, Giurgiu, Jastrzebski,
  Morrone, de~Laroussilhe, Gesmundo, Attariyan, and
  Gelly}]{pmlr-v97-houlsby19a}
Neil Houlsby, Andrei Giurgiu, Stanislaw Jastrzebski, Bruna Morrone, Quentin
  de~Laroussilhe, Andrea Gesmundo, Mona Attariyan, and Sylvain Gelly.
  2019{\natexlab{a}}.
\newblock \href {http://proceedings.mlr.press/v97/houlsby19a.html}
  {Parameter-efficient transfer learning for {NLP}}.
\newblock In \emph{Proceedings of the 36th International Conference on Machine
  Learning, {ICML} 2019, 9-15 June 2019, Long Beach, California, {USA}},
  volume~97 of \emph{Proceedings of Machine Learning Research}, pages
  2790--2799. {PMLR}.

\bibitem[{Houlsby et~al.(2019{\natexlab{b}})Houlsby, Giurgiu, Jastrzebski,
  Morrone, de~Laroussilhe, Gesmundo, Attariyan, and
  Gelly}]{houlsby2019parameter}
Neil Houlsby, Andrei Giurgiu, Stanislaw Jastrzebski, Bruna Morrone, Quentin
  de~Laroussilhe, Andrea Gesmundo, Mona Attariyan, and Sylvain Gelly.
  2019{\natexlab{b}}.
\newblock \href {http://proceedings.mlr.press/v97/houlsby19a.html}
  {Parameter-efficient transfer learning for {NLP}}.
\newblock In \emph{Proceedings of the 36th International Conference on Machine
  Learning, {ICML} 2019, 9-15 June 2019, Long Beach, California, {USA}},
  volume~97 of \emph{Proceedings of Machine Learning Research}, pages
  2790--2799. {PMLR}.

\bibitem[{Hu et~al.(2022{\natexlab{a}})Hu, Shen, Wallis, Allen{-}Zhu, Li, Wang,
  Wang, and Chen}]{hu2022lora}
Edward~J. Hu, Yelong Shen, Phillip Wallis, Zeyuan Allen{-}Zhu, Yuanzhi Li,
  Shean Wang, Lu~Wang, and Weizhu Chen. 2022{\natexlab{a}}.
\newblock \href {https://openreview.net/forum?id=nZeVKeeFYf9} {Lora: Low-rank
  adaptation of large language models}.
\newblock In \emph{The Tenth International Conference on Learning
  Representations, {ICLR} 2022, Virtual Event, April 25-29, 2022}.
  OpenReview.net.

\bibitem[{Hu et~al.(2022{\natexlab{b}})Hu, Shen, Wallis, Allen{-}Zhu, Li, Wang,
  Wang, and Chen}]{hu2021lora}
Edward~J. Hu, Yelong Shen, Phillip Wallis, Zeyuan Allen{-}Zhu, Yuanzhi Li,
  Shean Wang, Lu~Wang, and Weizhu Chen. 2022{\natexlab{b}}.
\newblock \href {https://openreview.net/forum?id=nZeVKeeFYf9} {Lora: Low-rank
  adaptation of large language models}.
\newblock In \emph{The Tenth International Conference on Learning
  Representations, {ICLR} 2022, Virtual Event, April 25-29, 2022}.
  OpenReview.net.

\bibitem[{Hu et~al.(2022{\natexlab{c}})Hu, Zhang, Ding, Wang, Wang, Liu, and
  Sun}]{hu2022sparse}
Shengding Hu, Zhen Zhang, Ning Ding, Yadao Wang, Yasheng Wang, Zhiyuan Liu, and
  Maosong Sun. 2022{\natexlab{c}}.
\newblock \href {https://openreview.net/forum?id=oOte_397Q4P} {Sparse structure
  search for delta tuning}.
\newblock In \emph{Advances in Neural Information Processing Systems}.

\bibitem[{Lawton et~al.(2023)Lawton, Kumar, Thattai, Galstyan, and
  Ver~Steeg}]{lawton-etal-2023-neural}
Neal Lawton, Anoop Kumar, Govind Thattai, Aram Galstyan, and Greg Ver~Steeg.
  2023.
\newblock \href {https://doi.org/10.18653/v1/2023.findings-acl.539} {Neural
  architecture search for parameter-efficient fine-tuning of large pre-trained
  language models}.
\newblock In \emph{Findings of the Association for Computational Linguistics:
  ACL 2023}, pages 8506--8515, Toronto, Canada. Association for Computational
  Linguistics.

\bibitem[{Lee et~al.(2019)Lee, Chang, and Toutanova}]{lee-etal-2019-latent}
Kenton Lee, Ming-Wei Chang, and Kristina Toutanova. 2019.
\newblock \href {https://doi.org/10.18653/v1/P19-1612} {Latent retrieval for
  weakly supervised open domain question answering}.
\newblock In \emph{Proceedings of the 57th Annual Meeting of the Association
  for Computational Linguistics}, pages 6086--6096, Florence, Italy.
  Association for Computational Linguistics.

\bibitem[{Lester et~al.(2021)Lester, Al-Rfou, and
  Constant}]{lester-etal-2021-power}
Brian Lester, Rami Al-Rfou, and Noah Constant. 2021.
\newblock \href {https://doi.org/10.18653/v1/2021.emnlp-main.243} {The power of
  scale for parameter-efficient prompt tuning}.
\newblock In \emph{Proceedings of the 2021 Conference on Empirical Methods in
  Natural Language Processing}, pages 3045--3059, Online and Punta Cana,
  Dominican Republic. Association for Computational Linguistics.

\bibitem[{Li and Liang(2021)}]{li-liang-2021-prefix}
Xiang~Lisa Li and Percy Liang. 2021.
\newblock \href {https://doi.org/10.18653/v1/2021.acl-long.353} {Prefix-tuning:
  Optimizing continuous prompts for generation}.
\newblock In \emph{Proceedings of the 59th Annual Meeting of the Association
  for Computational Linguistics and the 11th International Joint Conference on
  Natural Language Processing (Volume 1: Long Papers)}, pages 4582--4597,
  Online. Association for Computational Linguistics.

\bibitem[{Maas et~al.(2011)Maas, Daly, Pham, Huang, Ng, and
  Potts}]{maas-EtAl:2011:ACL-HLT2011}
Andrew~L. Maas, Raymond~E. Daly, Peter~T. Pham, Dan Huang, Andrew~Y. Ng, and
  Christopher Potts. 2011.
\newblock \href {https://aclanthology.org/P11-1015} {Learning word vectors for
  sentiment analysis}.
\newblock In \emph{Proceedings of the 49th Annual Meeting of the Association
  for Computational Linguistics: Human Language Technologies}, pages 142--150,
  Portland, Oregon, USA. Association for Computational Linguistics.

\bibitem[{Mahabadi et~al.(2021{\natexlab{a}})Mahabadi, Henderson, and
  Ruder}]{NEURIPS2021_081be9fd}
Rabeeh~Karimi Mahabadi, James Henderson, and Sebastian Ruder.
  2021{\natexlab{a}}.
\newblock \href
  {https://proceedings.neurips.cc/paper/2021/hash/081be9fdff07f3bc808f935906ef70c0-Abstract.html}
  {Compacter: Efficient low-rank hypercomplex adapter layers}.
\newblock In \emph{Advances in Neural Information Processing Systems 34: Annual
  Conference on Neural Information Processing Systems 2021, NeurIPS 2021,
  December 6-14, 2021, virtual}, pages 1022--1035.

\bibitem[{Mahabadi et~al.(2021{\natexlab{b}})Mahabadi, Henderson, and
  Ruder}]{compacter2021}
Rabeeh~Karimi Mahabadi, James Henderson, and Sebastian Ruder.
  2021{\natexlab{b}}.
\newblock \href
  {https://proceedings.neurips.cc/paper/2021/hash/081be9fdff07f3bc808f935906ef70c0-Abstract.html}
  {Compacter: Efficient low-rank hypercomplex adapter layers}.
\newblock In \emph{Advances in Neural Information Processing Systems 34: Annual
  Conference on Neural Information Processing Systems 2021, NeurIPS 2021,
  December 6-14, 2021, virtual}, pages 1022--1035.

\bibitem[{Mahabadi et~al.(2021{\natexlab{c}})Mahabadi, Henderson, and
  Ruder}]{karimi2021compacter}
Rabeeh~Karimi Mahabadi, James Henderson, and Sebastian Ruder.
  2021{\natexlab{c}}.
\newblock \href
  {https://proceedings.neurips.cc/paper/2021/hash/081be9fdff07f3bc808f935906ef70c0-Abstract.html}
  {Compacter: Efficient low-rank hypercomplex adapter layers}.
\newblock In \emph{Advances in Neural Information Processing Systems 34: Annual
  Conference on Neural Information Processing Systems 2021, NeurIPS 2021,
  December 6-14, 2021, virtual}, pages 1022--1035.

\bibitem[{Pang and Lee(2005)}]{pang-lee-2005-seeing}
Bo~Pang and Lillian Lee. 2005.
\newblock \href {https://doi.org/10.3115/1219840.1219855} {Seeing stars:
  Exploiting class relationships for sentiment categorization with respect to
  rating scales}.
\newblock In \emph{Proceedings of the 43rd Annual Meeting of the Association
  for Computational Linguistics ({ACL}{'}05)}, pages 115--124, Ann Arbor,
  Michigan. Association for Computational Linguistics.

\bibitem[{Pfeiffer et~al.(2021)Pfeiffer, Kamath, R{\"u}ckl{\'e}, Cho, and
  Gurevych}]{pfeiffer-etal-2021-adapterfusion}
Jonas Pfeiffer, Aishwarya Kamath, Andreas R{\"u}ckl{\'e}, Kyunghyun Cho, and
  Iryna Gurevych. 2021.
\newblock \href {https://doi.org/10.18653/v1/2021.eacl-main.39}
  {{A}dapter{F}usion: Non-destructive task composition for transfer learning}.
\newblock In \emph{Proceedings of the 16th Conference of the European Chapter
  of the Association for Computational Linguistics: Main Volume}, pages
  487--503, Online. Association for Computational Linguistics.

\bibitem[{Prasanna et~al.(2020)Prasanna, Rogers, and
  Rumshisky}]{prasanna-etal-2020-bert}
Sai Prasanna, Anna Rogers, and Anna Rumshisky. 2020.
\newblock \href {https://doi.org/10.18653/v1/2020.emnlp-main.259} {{W}hen
  {BERT} {P}lays the {L}ottery, {A}ll {T}ickets {A}re {W}inning}.
\newblock In \emph{Proceedings of the 2020 Conference on Empirical Methods in
  Natural Language Processing (EMNLP)}, pages 3208--3229, Online. Association
  for Computational Linguistics.

\bibitem[{Qin et~al.(2021)Qin, Wang, Su, Lin, Ding, Liu, Li, Hou, Li, Sun
  et~al.}]{qin2021exploring}
Yujia Qin, Xiaozhi Wang, Yusheng Su, Yankai Lin, Ning Ding, Zhiyuan Liu, Juanzi
  Li, Lei Hou, Peng Li, Maosong Sun, et~al. 2021.
\newblock \href {https://arxiv.org/abs/2110.07867} {Exploring low-dimensional
  intrinsic task subspace via prompt tuning}.
\newblock \emph{ArXiv preprint}, abs/2110.07867.

\bibitem[{Radford et~al.(2018)Radford, Narasimhan, Salimans, and
  Sutskever}]{radford2018improving}
Alec Radford, Karthik Narasimhan, Tim Salimans, and Ilya Sutskever. 2018.
\newblock \href
  {https://www.cs.ubc.ca/~amuham01/LING530/papers/radford2018improving.pdf}
  {Improving language understanding by generative pre-training}.

\bibitem[{Raffel et~al.(2020)Raffel, Shazeer, Roberts, Lee, Narang, Matena,
  Zhou, Li, and Liu}]{raffel2020exploring}
Colin Raffel, Noam Shazeer, Adam Roberts, Katherine Lee, Sharan Narang, Michael
  Matena, Yanqi Zhou, Wei Li, and Peter~J. Liu. 2020.
\newblock \href {http://jmlr.org/papers/v21/20-074.html} {Exploring the limits
  of transfer learning with a unified text-to-text transformer}.
\newblock \emph{J. Mach. Learn. Res.}, 21:140:1--140:67.

\bibitem[{Scao et~al.(2023)Scao, Fan, Akiki, Pavlick, Ilić, Hesslow,
  Castagné, Luccioni, Yvon, Gallé, Tow, Rush, Biderman, Webson, Ammanamanchi,
  Wang, Sagot, Muennighoff, del Moral, Ruwase, Bawden, Bekman, McMillan-Major,
  Beltagy, Nguyen, Saulnier, Tan, Suarez, Sanh, Laurençon, Jernite, Launay,
  Mitchell, Raffel, Gokaslan, Simhi, Soroa, Aji, Alfassy, Rogers, Nitzav, Xu,
  Mou, Emezue, Klamm, Leong, van Strien, Adelani, Radev, Ponferrada, Levkovizh,
  Kim, Natan, Toni, Dupont, Kruszewski, Pistilli, Elsahar, Benyamina, Tran, Yu,
  Abdulmumin, Johnson, Gonzalez-Dios, de~la Rosa, Chim, Dodge, Zhu, Chang,
  Frohberg, Tobing, Bhattacharjee, Almubarak, Chen, Lo, Werra, Weber, Phan,
  allal, Tanguy, Dey, Muñoz, Masoud, Grandury, Šaško, Huang, Coavoux, Singh,
  Jiang, Vu, Jauhar, Ghaleb, Subramani, Kassner, Khamis, Nguyen, Espejel,
  de~Gibert, Villegas, Henderson, Colombo, Amuok, Lhoest, Harliman, Bommasani,
  López, Ribeiro, Osei, Pyysalo, Nagel, Bose, Muhammad, Sharma, Longpre,
  Nikpoor, Silberberg, Pai, Zink, Torrent, Schick, Thrush, Danchev, Nikoulina,
  Laippala, Lepercq, Prabhu, Alyafeai, Talat, Raja, Heinzerling, Si, Taşar,
  Salesky, Mielke, Lee, Sharma, Santilli, Chaffin, Stiegler, Datta, Szczechla,
  Chhablani, Wang, Pandey, Strobelt, Fries, Rozen, Gao, Sutawika, Bari,
  Al-shaibani, Manica, Nayak, Teehan, Albanie, Shen, Ben-David, Bach, Kim,
  Bers, Fevry, Neeraj, Thakker, Raunak, Tang, Yong, Sun, Brody, Uri, Tojarieh,
  Roberts, Chung, Tae, Phang, Press, Li, Narayanan, Bourfoune, Casper, Rasley,
  Ryabinin, Mishra, Zhang, Shoeybi, Peyrounette, Patry, Tazi, Sanseviero, von
  Platen, Cornette, Lavallée, Lacroix, Rajbhandari, Gandhi, Smith, Requena,
  Patil, Dettmers, Baruwa, Singh, Cheveleva, Ligozat, Subramonian, Névéol,
  Lovering, Garrette, Tunuguntla, Reiter, Taktasheva, Voloshina, Bogdanov,
  Winata, Schoelkopf, Kalo, Novikova, Forde, Clive, Kasai, Kawamura, Hazan,
  Carpuat, Clinciu, Kim, Cheng, Serikov, Antverg, van~der Wal, Zhang, Zhang,
  Gehrmann, Mirkin, Pais, Shavrina, Scialom, Yun, Limisiewicz, Rieser,
  Protasov, Mikhailov, Pruksachatkun, Belinkov, Bamberger, Kasner, Rueda,
  Pestana, Feizpour, Khan, Faranak, Santos, Hevia, Unldreaj, Aghagol,
  Abdollahi, Tammour, HajiHosseini, Behroozi, Ajibade, Saxena, Ferrandis,
  Contractor, Lansky, David, Kiela, Nguyen, Tan, Baylor, Ozoani, Mirza,
  Ononiwu, Rezanejad, Jones, Bhattacharya, Solaiman, Sedenko, Nejadgholi,
  Passmore, Seltzer, Sanz, Dutra, Samagaio, Elbadri, Mieskes, Gerchick,
  Akinlolu, McKenna, Qiu, Ghauri, Burynok, Abrar, Rajani, Elkott, Fahmy,
  Samuel, An, Kromann, Hao, Alizadeh, Shubber, Wang, Roy, Viguier, Le, Oyebade,
  Le, Yang, Nguyen, Kashyap, Palasciano, Callahan, Shukla, Miranda-Escalada,
  Singh, Beilharz, Wang, Brito, Zhou, Jain, Xu, Fourrier, Periñán, Molano,
  Yu, Manjavacas, Barth, Fuhrimann, Altay, Bayrak, Burns, Vrabec, Bello, Dash,
  Kang, Giorgi, Golde, Posada, Sivaraman, Bulchandani, Liu, Shinzato,
  de~Bykhovetz, Takeuchi, Pàmies, Castillo, Nezhurina, Sänger, Samwald,
  Cullan, Weinberg, Wolf, Mihaljcic, Liu, Freidank, Kang, Seelam, Dahlberg,
  Broad, Muellner, Fung, Haller, Chandrasekhar, Eisenberg, Martin, Canalli, Su,
  Su, Cahyawijaya, Garda, Deshmukh, Mishra, Kiblawi, Ott, Sang-aroonsiri,
  Kumar, Schweter, Bharati, Laud, Gigant, Kainuma, Kusa, Labrak, Bajaj,
  Venkatraman, Xu, Xu, Xu, Tan, Xie, Ye, Bras, Belkada, and Wolf}]{2023bloom}
Teven~Le Scao, Angela Fan, Christopher Akiki, Ellie Pavlick, Suzana Ilić,
  Daniel Hesslow, Roman Castagné, Alexandra~Sasha Luccioni, François Yvon,
  Matthias Gallé, Jonathan Tow, Alexander~M. Rush, Stella Biderman, Albert
  Webson, Pawan~Sasanka Ammanamanchi, Thomas Wang, Benoît Sagot, Niklas
  Muennighoff, Albert~Villanova del Moral, Olatunji Ruwase, Rachel Bawden, Stas
  Bekman, Angelina McMillan-Major, Iz~Beltagy, Huu Nguyen, Lucile Saulnier,
  Samson Tan, Pedro~Ortiz Suarez, Victor Sanh, Hugo Laurençon, Yacine Jernite,
  Julien Launay, Margaret Mitchell, Colin Raffel, Aaron Gokaslan, Adi Simhi,
  Aitor Soroa, Alham~Fikri Aji, Amit Alfassy, Anna Rogers, Ariel~Kreisberg
  Nitzav, Canwen Xu, Chenghao Mou, Chris Emezue, Christopher Klamm, Colin
  Leong, Daniel van Strien, David~Ifeoluwa Adelani, Dragomir Radev,
  Eduardo~González Ponferrada, Efrat Levkovizh, Ethan Kim, Eyal~Bar Natan,
  Francesco~De Toni, Gérard Dupont, Germán Kruszewski, Giada Pistilli, Hady
  Elsahar, Hamza Benyamina, Hieu Tran, Ian Yu, Idris Abdulmumin, Isaac Johnson,
  Itziar Gonzalez-Dios, Javier de~la Rosa, Jenny Chim, Jesse Dodge, Jian Zhu,
  Jonathan Chang, Jörg Frohberg, Joseph Tobing, Joydeep Bhattacharjee, Khalid
  Almubarak, Kimbo Chen, Kyle Lo, Leandro~Von Werra, Leon Weber, Long Phan,
  Loubna~Ben allal, Ludovic Tanguy, Manan Dey, Manuel~Romero Muñoz, Maraim
  Masoud, María Grandury, Mario Šaško, Max Huang, Maximin Coavoux, Mayank
  Singh, Mike Tian-Jian Jiang, Minh~Chien Vu, Mohammad~A. Jauhar, Mustafa
  Ghaleb, Nishant Subramani, Nora Kassner, Nurulaqilla Khamis, Olivier Nguyen,
  Omar Espejel, Ona de~Gibert, Paulo Villegas, Peter Henderson, Pierre Colombo,
  Priscilla Amuok, Quentin Lhoest, Rheza Harliman, Rishi Bommasani,
  Roberto~Luis López, Rui Ribeiro, Salomey Osei, Sampo Pyysalo, Sebastian
  Nagel, Shamik Bose, Shamsuddeen~Hassan Muhammad, Shanya Sharma, Shayne
  Longpre, Somaieh Nikpoor, Stanislav Silberberg, Suhas Pai, Sydney Zink,
  Tiago~Timponi Torrent, Timo Schick, Tristan Thrush, Valentin Danchev,
  Vassilina Nikoulina, Veronika Laippala, Violette Lepercq, Vrinda Prabhu, Zaid
  Alyafeai, Zeerak Talat, Arun Raja, Benjamin Heinzerling, Chenglei Si,
  Davut~Emre Taşar, Elizabeth Salesky, Sabrina~J. Mielke, Wilson~Y. Lee,
  Abheesht Sharma, Andrea Santilli, Antoine Chaffin, Arnaud Stiegler, Debajyoti
  Datta, Eliza Szczechla, Gunjan Chhablani, Han Wang, Harshit Pandey, Hendrik
  Strobelt, Jason~Alan Fries, Jos Rozen, Leo Gao, Lintang Sutawika, M~Saiful
  Bari, Maged~S. Al-shaibani, Matteo Manica, Nihal Nayak, Ryan Teehan, Samuel
  Albanie, Sheng Shen, Srulik Ben-David, Stephen~H. Bach, Taewoon Kim, Tali
  Bers, Thibault Fevry, Trishala Neeraj, Urmish Thakker, Vikas Raunak, Xiangru
  Tang, Zheng-Xin Yong, Zhiqing Sun, Shaked Brody, Yallow Uri, Hadar Tojarieh,
  Adam Roberts, Hyung~Won Chung, Jaesung Tae, Jason Phang, Ofir Press, Conglong
  Li, Deepak Narayanan, Hatim Bourfoune, Jared Casper, Jeff Rasley, Max
  Ryabinin, Mayank Mishra, Minjia Zhang, Mohammad Shoeybi, Myriam Peyrounette,
  Nicolas Patry, Nouamane Tazi, Omar Sanseviero, Patrick von Platen, Pierre
  Cornette, Pierre~François Lavallée, Rémi Lacroix, Samyam Rajbhandari,
  Sanchit Gandhi, Shaden Smith, Stéphane Requena, Suraj Patil, Tim Dettmers,
  Ahmed Baruwa, Amanpreet Singh, Anastasia Cheveleva, Anne-Laure Ligozat, Arjun
  Subramonian, Aurélie Névéol, Charles Lovering, Dan Garrette, Deepak
  Tunuguntla, Ehud Reiter, Ekaterina Taktasheva, Ekaterina Voloshina, Eli
  Bogdanov, Genta~Indra Winata, Hailey Schoelkopf, Jan-Christoph Kalo,
  Jekaterina Novikova, Jessica~Zosa Forde, Jordan Clive, Jungo Kasai, Ken
  Kawamura, Liam Hazan, Marine Carpuat, Miruna Clinciu, Najoung Kim, Newton
  Cheng, Oleg Serikov, Omer Antverg, Oskar van~der Wal, Rui Zhang, Ruochen
  Zhang, Sebastian Gehrmann, Shachar Mirkin, Shani Pais, Tatiana Shavrina,
  Thomas Scialom, Tian Yun, Tomasz Limisiewicz, Verena Rieser, Vitaly Protasov,
  Vladislav Mikhailov, Yada Pruksachatkun, Yonatan Belinkov, Zachary Bamberger,
  Zdeněk Kasner, Alice Rueda, Amanda Pestana, Amir Feizpour, Ammar Khan, Amy
  Faranak, Ana Santos, Anthony Hevia, Antigona Unldreaj, Arash Aghagol, Arezoo
  Abdollahi, Aycha Tammour, Azadeh HajiHosseini, Bahareh Behroozi, Benjamin
  Ajibade, Bharat Saxena, Carlos~Muñoz Ferrandis, Danish Contractor, David
  Lansky, Davis David, Douwe Kiela, Duong~A. Nguyen, Edward Tan, Emi Baylor,
  Ezinwanne Ozoani, Fatima Mirza, Frankline Ononiwu, Habib Rezanejad, Hessie
  Jones, Indrani Bhattacharya, Irene Solaiman, Irina Sedenko, Isar Nejadgholi,
  Jesse Passmore, Josh Seltzer, Julio~Bonis Sanz, Livia Dutra, Mairon Samagaio,
  Maraim Elbadri, Margot Mieskes, Marissa Gerchick, Martha Akinlolu, Michael
  McKenna, Mike Qiu, Muhammed Ghauri, Mykola Burynok, Nafis Abrar, Nazneen
  Rajani, Nour Elkott, Nour Fahmy, Olanrewaju Samuel, Ran An, Rasmus Kromann,
  Ryan Hao, Samira Alizadeh, Sarmad Shubber, Silas Wang, Sourav Roy, Sylvain
  Viguier, Thanh Le, Tobi Oyebade, Trieu Le, Yoyo Yang, Zach Nguyen,
  Abhinav~Ramesh Kashyap, Alfredo Palasciano, Alison Callahan, Anima Shukla,
  Antonio Miranda-Escalada, Ayush Singh, Benjamin Beilharz, Bo~Wang, Caio
  Brito, Chenxi Zhou, Chirag Jain, Chuxin Xu, Clémentine Fourrier,
  Daniel~León Periñán, Daniel Molano, Dian Yu, Enrique Manjavacas, Fabio
  Barth, Florian Fuhrimann, Gabriel Altay, Giyaseddin Bayrak, Gully Burns,
  Helena~U. Vrabec, Imane Bello, Ishani Dash, Jihyun Kang, John Giorgi, Jonas
  Golde, Jose~David Posada, Karthik~Rangasai Sivaraman, Lokesh Bulchandani,
  Lu~Liu, Luisa Shinzato, Madeleine~Hahn de~Bykhovetz, Maiko Takeuchi, Marc
  Pàmies, Maria~A Castillo, Marianna Nezhurina, Mario Sänger, Matthias
  Samwald, Michael Cullan, Michael Weinberg, Michiel~De Wolf, Mina Mihaljcic,
  Minna Liu, Moritz Freidank, Myungsun Kang, Natasha Seelam, Nathan Dahlberg,
  Nicholas~Michio Broad, Nikolaus Muellner, Pascale Fung, Patrick Haller, Ramya
  Chandrasekhar, Renata Eisenberg, Robert Martin, Rodrigo Canalli, Rosaline Su,
  Ruisi Su, Samuel Cahyawijaya, Samuele Garda, Shlok~S Deshmukh, Shubhanshu
  Mishra, Sid Kiblawi, Simon Ott, Sinee Sang-aroonsiri, Srishti Kumar, Stefan
  Schweter, Sushil Bharati, Tanmay Laud, Théo Gigant, Tomoya Kainuma, Wojciech
  Kusa, Yanis Labrak, Yash~Shailesh Bajaj, Yash Venkatraman, Yifan Xu, Yingxin
  Xu, Yu~Xu, Zhe Tan, Zhongli Xie, Zifan Ye, Mathilde Bras, Younes Belkada, and
  Thomas Wolf. 2023.
\newblock \href {http://arxiv.org/abs/2211.05100} {Bloom: A 176b-parameter
  open-access multilingual language model}.

\bibitem[{Sharma et~al.(2019)Sharma, Graesser, Nangia, and
  Evci}]{sharma2019natural}
Lakshay Sharma, Laura Graesser, Nikita Nangia, and Utku Evci. 2019.
\newblock \href {https://arxiv.org/abs/1907.01041} {Natural language
  understanding with the quora question pairs dataset}.
\newblock \emph{ArXiv preprint}, abs/1907.01041.

\bibitem[{Socher et~al.(2013)Socher, Perelygin, Wu, Chuang, Manning, Ng, and
  Potts}]{socher-etal-2013-recursive}
Richard Socher, Alex Perelygin, Jean Wu, Jason Chuang, Christopher~D. Manning,
  Andrew Ng, and Christopher Potts. 2013.
\newblock \href {https://aclanthology.org/D13-1170} {Recursive deep models for
  semantic compositionality over a sentiment treebank}.
\newblock In \emph{Proceedings of the 2013 Conference on Empirical Methods in
  Natural Language Processing}, pages 1631--1642, Seattle, Washington, USA.
  Association for Computational Linguistics.

\bibitem[{Su et~al.(2022)Su, Wang, Qin, Chan, Lin, Wang, Wen, Liu, Li, Li, Hou,
  Sun, and Zhou}]{transferabilitysu}
Yusheng Su, Xiaozhi Wang, Yujia Qin, Chi-Min Chan, Yankai Lin, Huadong Wang,
  Kaiyue Wen, Zhiyuan Liu, Peng Li, Juanzi Li, Lei Hou, Maosong Sun, and Jie
  Zhou. 2022.
\newblock \href {https://doi.org/10.18653/v1/2022.naacl-main.290} {On
  transferability of prompt tuning for natural language processing}.
\newblock In \emph{Proceedings of the 2022 Conference of the North American
  Chapter of the Association for Computational Linguistics: Human Language
  Technologies}, pages 3949--3969, Seattle, United States. Association for
  Computational Linguistics.

\bibitem[{Vaswani et~al.(2017)Vaswani, Shazeer, Parmar, Uszkoreit, Jones,
  Gomez, Kaiser, and Polosukhin}]{NIPS2017_3f5ee243}
Ashish Vaswani, Noam Shazeer, Niki Parmar, Jakob Uszkoreit, Llion Jones,
  Aidan~N. Gomez, Lukasz Kaiser, and Illia Polosukhin. 2017.
\newblock \href
  {https://proceedings.neurips.cc/paper/2017/hash/3f5ee243547dee91fbd053c1c4a845aa-Abstract.html}
  {Attention is all you need}.
\newblock In \emph{Advances in Neural Information Processing Systems 30: Annual
  Conference on Neural Information Processing Systems 2017, December 4-9, 2017,
  Long Beach, CA, {USA}}, pages 5998--6008.

\bibitem[{Vu et~al.(2022)Vu, Lester, Constant, Al-Rfou{'}, and
  Cer}]{vu-etal-2022-spot}
Tu~Vu, Brian Lester, Noah Constant, Rami Al-Rfou{'}, and Daniel Cer. 2022.
\newblock \href {https://doi.org/10.18653/v1/2022.acl-long.346} {{SP}o{T}:
  Better frozen model adaptation through soft prompt transfer}.
\newblock In \emph{Proceedings of the 60th Annual Meeting of the Association
  for Computational Linguistics (Volume 1: Long Papers)}, pages 5039--5059,
  Dublin, Ireland. Association for Computational Linguistics.

\bibitem[{Wang et~al.(2019)Wang, Singh, Michael, Hill, Levy, and
  Bowman}]{wang2018glue}
Alex Wang, Amanpreet Singh, Julian Michael, Felix Hill, Omer Levy, and
  Samuel~R. Bowman. 2019.
\newblock \href {https://openreview.net/forum?id=rJ4km2R5t7} {{GLUE:} {A}
  multi-task benchmark and analysis platform for natural language
  understanding}.
\newblock In \emph{7th International Conference on Learning Representations,
  {ICLR} 2019, New Orleans, LA, USA, May 6-9, 2019}. OpenReview.net.

\bibitem[{Wei et~al.(2022{\natexlab{a}})Wei, Tay, Bommasani, Raffel, Zoph,
  Borgeaud, Yogatama, Bosma, Zhou, Metzler, Chi, Hashimoto, Vinyals, Liang,
  Dean, and Fedus}]{wei2022emergent}
Jason Wei, Yi~Tay, Rishi Bommasani, Colin Raffel, Barret Zoph, Sebastian
  Borgeaud, Dani Yogatama, Maarten Bosma, Denny Zhou, Donald Metzler, Ed~H.
  Chi, Tatsunori Hashimoto, Oriol Vinyals, Percy Liang, Jeff Dean, and William
  Fedus. 2022{\natexlab{a}}.
\newblock \href {https://openreview.net/forum?id=yzkSU5zdwD} {Emergent
  abilities of large language models}.
\newblock \emph{Transactions on Machine Learning Research}.

\bibitem[{Wei et~al.(2022{\natexlab{b}})Wei, Wang, Schuurmans, Bosma, Ichter,
  Xia, Chi, Le, and Zhou}]{wei2022_chain_of_thought}
Jason Wei, Xuezhi Wang, Dale Schuurmans, Maarten Bosma, Brian Ichter, Fei Xia,
  Ed~Chi, Quoc Le, and Denny Zhou. 2022{\natexlab{b}}.
\newblock \href {https://arxiv.org/abs/2201.11903} {Chain of thought prompting
  elicits reasoning in large language models}.

\bibitem[{Williams et~al.(2018)Williams, Nangia, and
  Bowman}]{williams-etal-2018-broad}
Adina Williams, Nikita Nangia, and Samuel Bowman. 2018.
\newblock \href {https://doi.org/10.18653/v1/N18-1101} {A broad-coverage
  challenge corpus for sentence understanding through inference}.
\newblock In \emph{Proceedings of the 2018 Conference of the North {A}merican
  Chapter of the Association for Computational Linguistics: Human Language
  Technologies, Volume 1 (Long Papers)}, pages 1112--1122, New Orleans,
  Louisiana. Association for Computational Linguistics.

\bibitem[{Yang and Liu(2022)}]{yang2022on}
Zonghan Yang and Yang Liu. 2022.
\newblock \href {https://openreview.net/forum?id=eBCmOocUejf} {On robust
  prefix-tuning for text classification}.
\newblock In \emph{The Tenth International Conference on Learning
  Representations, {ICLR} 2022, Virtual Event, April 25-29, 2022}.
  OpenReview.net.

\bibitem[{Yi et~al.(2022)Yi, Chen, Qin, Lin, Ding, Han, Liu, Sun, and
  Zhou}]{yi2022}
Jing Yi, Weize Chen, Yujia Qin, Yankai Lin, Ning Ding, Xu~Han, Zhiyuan Liu,
  Maosong Sun, and Jie Zhou. 2022.
\newblock \href {https://arxiv.org/abs/2210.13311} {Different tunes played with
  equal skill: Exploring a unified optimization subspace for delta tuning}.

\bibitem[{Zhou et~al.(2023)Zhou, Wan, Vuli{\'c}, and
  Korhonen}]{zhou2023autopeft}
Han Zhou, Xingchen Wan, Ivan Vuli{\'c}, and Anna Korhonen. 2023.
\newblock \href {https://arxiv.org/abs/2301.12132} {Autopeft: Automatic
  configuration search for parameter-efficient fine-tuning}.
\newblock \emph{arXiv preprint arXiv:2301.12132}.

\end{thebibliography}
\bibliographystyle{acl_natbib}

\clearpage
\appendix


\section{Task and Dataset}
\label{appendix:task_and_dataset}
We use various NLP tasks to evaluate the \APET methods, which can be divided into the following $5$ categories:

\paragraph{Sentiment Analysis (SA)} SA tasks evaluate if a model can correctly predict the sentiment labels of an input sentence. In this paper, we choose SST-2 \cite{socher-etal-2013-recursive}, IMDB \cite{maas-EtAl:2011:ACL-HLT2011}, and Rotten Tomatoes \cite{pang-lee-2005-seeing}.

\paragraph{Natural Language Inference (NLI)} NLI tasks evaluate a model's ability to correctly classify if a hypothesis can be entailed or not given a premise. In this paper, we choose MNLI \cite{williams-etal-2018-broad}, QNLI \cite{wang2018glue}, and RTE \cite{bos-markert-2005-recognising}.

\paragraph{Paraphrase Identification (PI)} PI tasks evaluate if a model can correctly identify paraphrases, which means two sentences are identical in semantic meaning. In this paper, we choose MRPC \cite{dolan-brockett-2005-automatically}, and QQP \cite{sharma2019natural}.

\paragraph{Question Answering (QA)} QA tasks evaluate a model's ability to answer questions. Context may be present. In this paper, we choose NQ-Open \cite{lee-etal-2019-latent}, an open-world QA dataset without context.

\paragraph{Summarization (SUM)} SUM tasks evaluate a model's ability to summarize a long paragraph into a shorter abstract without loosing the semantics of the original text. In this paper, we choose SamSUM \cite{gliwa-etal-2019-samsum}, and Multi-News \cite{fabbri-etal-2019-multi} in our experiments.



\section{Parameter-efficient Tuning (\PET) Methods}
\label{appendix:pet}

Here, we first recap the PLM (transformer) layer. Then, we describe the detail and training configurations of the \PET methods shown in Figure~\ref{fig:the_power_of_scale}.


\subsection{Transformer Architecture}
\label{appendix:transformer}
A PLM is generally a stack of multiple Transformer layers, each composed of a multi-headed attention and a feed-forward network. The multi-headed attention contains $h$ attention heads working in parallel. Specifically, given an input $\mathbf{X} \in \mathbb{R}^{n \times d}$, the $i$-th attention head works as follows:
\begin{equation}
    \mathbf{h}_i = \texttt{softmax}(\frac{(\mathbf{X W}^i_q) (\mathbf{X W}^i_k)^T}{\sqrt{d / h}} (\mathbf{X W}^i_v)),
\end{equation}

\noindent where $n$ is sequence length, $d$ is the hidden dimension, $\mathbf{W}^i_q \in \mathbb{R}^{n \times d}$ is query, $\mathbf{W}^i_k \in \mathbb{R}^{n \times d}$ is key, and $\mathbf{W}^i_v \in \mathbb{R}^{n \times d}$ is value. The output from each attention head will be concatenated and further transformed by $\mathbf{W}_o \in \mathbb{R}^{d \times d}$ and be denoted as:
\begin{equation}
    \mathbf{h}_{\texttt{MHA}}  = \texttt{concat}(\mathbf{h}_1, \mathbf{h}_2, ..., \mathbf{h}_h) \mathbf{W}_o,
\end{equation}
where $\mathbf{h}_{\texttt{MHA}} \in \mathbb{R}^{n \times d}$ is the output hidden state of  multi-headed attention layer. After that, $\mathbf{h}$ will be fed into a two-layer feed-forward network
\begin{equation}
    \mathbf{h}_{\texttt{FFN}} = \sigma(\mathbf{h W}_1 + \mathbf{b}_1) \mathbf{W}_2 + \mathbf{b}_2,
\end{equation}
\noindent where $\mathbf{W}_1 \in \mathbb{R}^{d \times d_m}$, $\mathbf{W}_2 \in \mathbb{R}^{d_m \times d}$, $\mathbf{b}_1 \in \mathbb{R}^{d_m}$, $\mathbf{b}_2 \in \mathbb{R}^d$, and $d_m > d$ is an integer.

During the forward pass through each (transformer) block, the input hidden state is applied with the sequence of layers. For simplicity, we formalize the transformation of each layer as 
\begin{equation}
    \mathbf{h}^{out}=f(\mathbf{h}^{in}).
\end{equation}
Under the layer as the operator $f$, the input hidden state $\mathbf{h}^{in}\in\mathbb{R}^{n \times d}$ is transformed into the output hidden state $\mathbf{h}^{out}\in\mathbb{R}^{n \times d}$, where $s$ is the input length, and $d$ is the dimension.

\subsection{Implementation Details of \PET Methods}



\paragraph{Prompt} Prompt-tuning \cite{lester-etal-2021-power} prepends $N_p$ tunable soft tokens, i.e. embeddings, to the input sentences and asks the model to predict the probability of the next word. During training, only the newly added embeddings are optimized and the backbone model is frozen. 

\paragraph{BitFit} BitFit \cite{ben-zaken-etal-2022-bitfit} is a method that only tunes all the bias terms $\mathbf{W}_b \in \mathbb{R}^{d}$ in the PLM, which lie in the self-attention and layer norm layers. 


\paragraph{LoRA} LoRA \cite{hu2021lora} is a method that adapts a PLM in a low-rank space. It down-projects the attention weights into a lower dimension and up-projects it back to the original dimension. Only these projection weights are optimized. 



\paragraph{Adapter} Adapter \cite{houlsby2019parameter} is a method that only tunes the inserted adapter modules, which consist of down projection, non-linear transformation, up projection, and a skip-connection. For each existing Transformer layer in a PLM, the adapter modules are inserted at two locations: (1) after the first feed-forward layer, and (2) after the two consecutive feed-forward layers. During training, only the adapter modules are optimized and the rest of the PLM is frozen. 




\subsection{Training Configurations of \PET Methods}
\label{appendix:training_configurations_of_pet_methods}

The tunable module of a \PET method $\theta$ is composed of $L$ tunable weights $\mathbf{W}$ (all tunable weights) of the specific \PET method, which can be expressed as $\theta= \{ \mathbf{{W}}_{1}, \mathbf{{W}}_{2}, ..., \mathbf{{W}}_{L} \}$. We also follow Equation \eqref{eq:pet_training_objective} to train the \PET  method. During training, we only optimize $\theta$ while freezing the rest of the parameters in the PLM. We adopt a batch size of $32$ and have no warm-up for most of the \PET models and tasks. The maximum input length is $128$ for single sentence tasks (SA) and $256$ for multi-sentence tasks (NLI, PI, QA, SUM). The maximum generation length is $1$ for classification tasks (SA, NLI, PI), $64$ for Multi-News, and $128$ for SAMSum. On the BERT, BLOOM, T5 models, we set their learning rates as \{3e-4\}, \{3e-4, 5e-5\}, \{1e-4, 1e-3, 1e-2\} respectively. Then, we choose the best performance to report.




\begin{figure}[!t]
\centering
\includegraphics[width=0.485\textwidth]{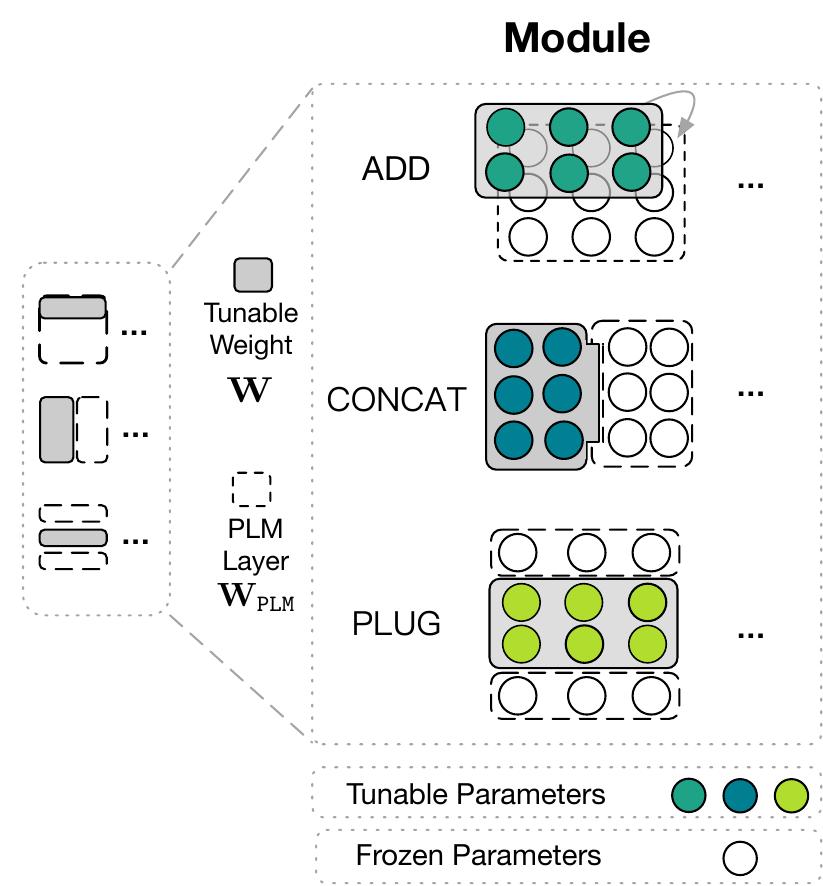}
\caption{The tunable modules of \APET methods are composed $p$ tunable weights $\mathbf{W}$, which can be expressed as $\theta= \{ \mathbf{W}_{1}, \mathbf{W}_{2}, ..., \mathbf{W}_{L} \}$. We introduce three  operations to insert the tunable weight $\mathbf{W}$ into the PLM and the corresponding transformation.}
\label{fig:appendix_afp}
\end{figure}

\section{Arbitrary Parameter-Efficient Tuning (\APET) Methods}
\label{appendix:appendix:afp}
We introduce a more flexible \PET method, \textbf{A}rbitrary \textbf{P}arameter-\textbf{E}fficient Tuning (\APET) method. Its tunable module can be arbitrary structure that facilitates us to explore various module structures (parameter position) and easier control the number of tunable parameters.


\subsection{Implementation Details of \APET Methods}
\label{}
As we previously introduced in \cref{ssec:afp_tuning_methods}, the tunable module of the \APET method is composed of tunable weights. Each tunable weight can be expressed as $\mathbf{W}$. Here, we have three operations to insert the tunable weight $\mathbf{W}$ into the PLM to modify the specific layers and their corresponding transformations as follows:

\paragraph{ADD}
\label{}
We will add the tunable weight $\mathbf{W}$ into the PLM layer. The corresponding transformation can be denoted as $\mathbf{h}^{out}$:
\begin{equation}
\label{appendix:afp_add}
    f(\mathbf{h}^{in}) + \mathbf{W}_{\texttt{1}}.
\end{equation}

\paragraph{CONCAT}
\label{}
We will concatenate the tunable weight $\mathbf{W}$ and the hidden state or the layer in the PLM. The corresponding transformation can be denoted as $\mathbf{h}^{out}$:
\begin{equation}
\label{appendix:afp_concatenate}
\begin{aligned}
     f(\mathbf{h}^{in}) +
     \left\{\begin{matrix}
     f(\mathbf{W}_{\texttt{2}})\\ 
     \alpha \mathbf{h}^{in} \mathbf{W}_{\texttt{3}} \mathbf{W}_{\texttt{4}} 
    \end{matrix}\right.
\end{aligned}
\end{equation}

\paragraph{PLUG}
\label{}
We will plug the tunable weight $\mathbf{W}$ between PLM layers. The corresponding transformation can be denoted as $\mathbf{h}^{out}$:
\begin{equation}
\label{appendix:afp_plugin}
    f(\mathbf{h}^{in}) + \sigma(f(\mathbf{h}^{in}) \mathbf{W}_{\texttt{5}} \mathbf{W}_{\texttt{6}}).
\end{equation}

According to these operations and the corresponding transformations, we can express the \APET methods as $\mathbf{h}^{out}$:
\begin{equation}
\label{appendix:afp_complete}
\begin{aligned}
     f(\mathbf{h}^{in}) + 
     \left\{\begin{matrix}
     \mathbf{W}_{\texttt{1}}\\
     f(\mathbf{W}_{\texttt{2}}) \\ 
     \alpha \mathbf{h}^{in} \mathbf{W}_{\texttt{3}}  \mathbf{W}_{\texttt{4}} \\
     \sigma(f(\mathbf{h}^{in}) \mathbf{W}_{\texttt{5}}) \mathbf{W}_{\texttt{6}}\\
     \vdots
    \end{matrix}\right.
\end{aligned}
\end{equation}
By comparing the Equation (\ref{appendix:afp_complete}) with the equations of the previously introduced \PET methods, we can clearly find that the \PET methods are special cases of \APET methods.

\begin{figure*}[!t]
    \centering
    \subfigure[\BERTSMALL and \TSMALL]{
	    \includegraphics[width=0.482\textwidth]{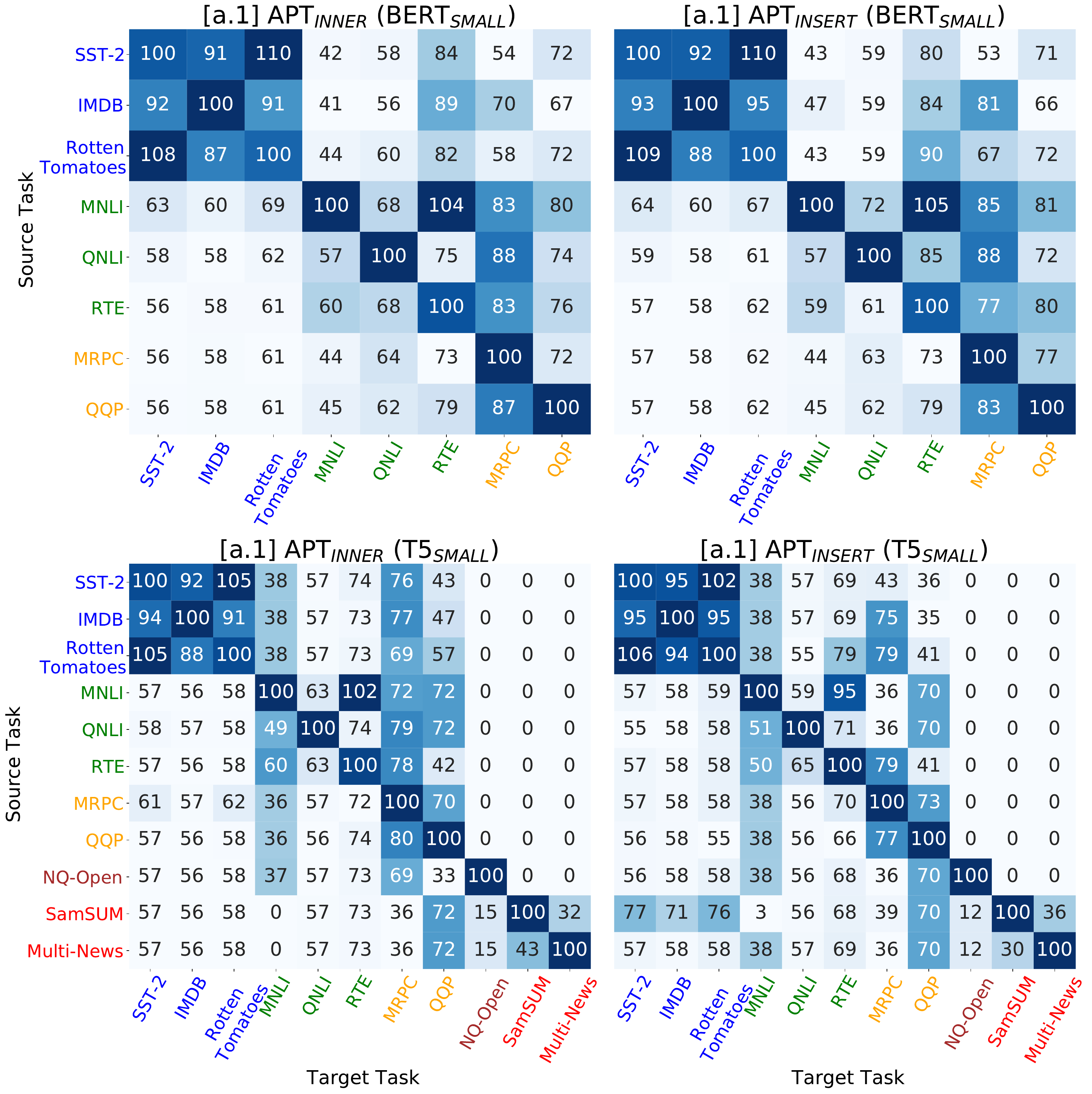}
    }
    \subfigure[\BERTLARGE and \TXXL]{
	    \includegraphics[width=0.482\textwidth]{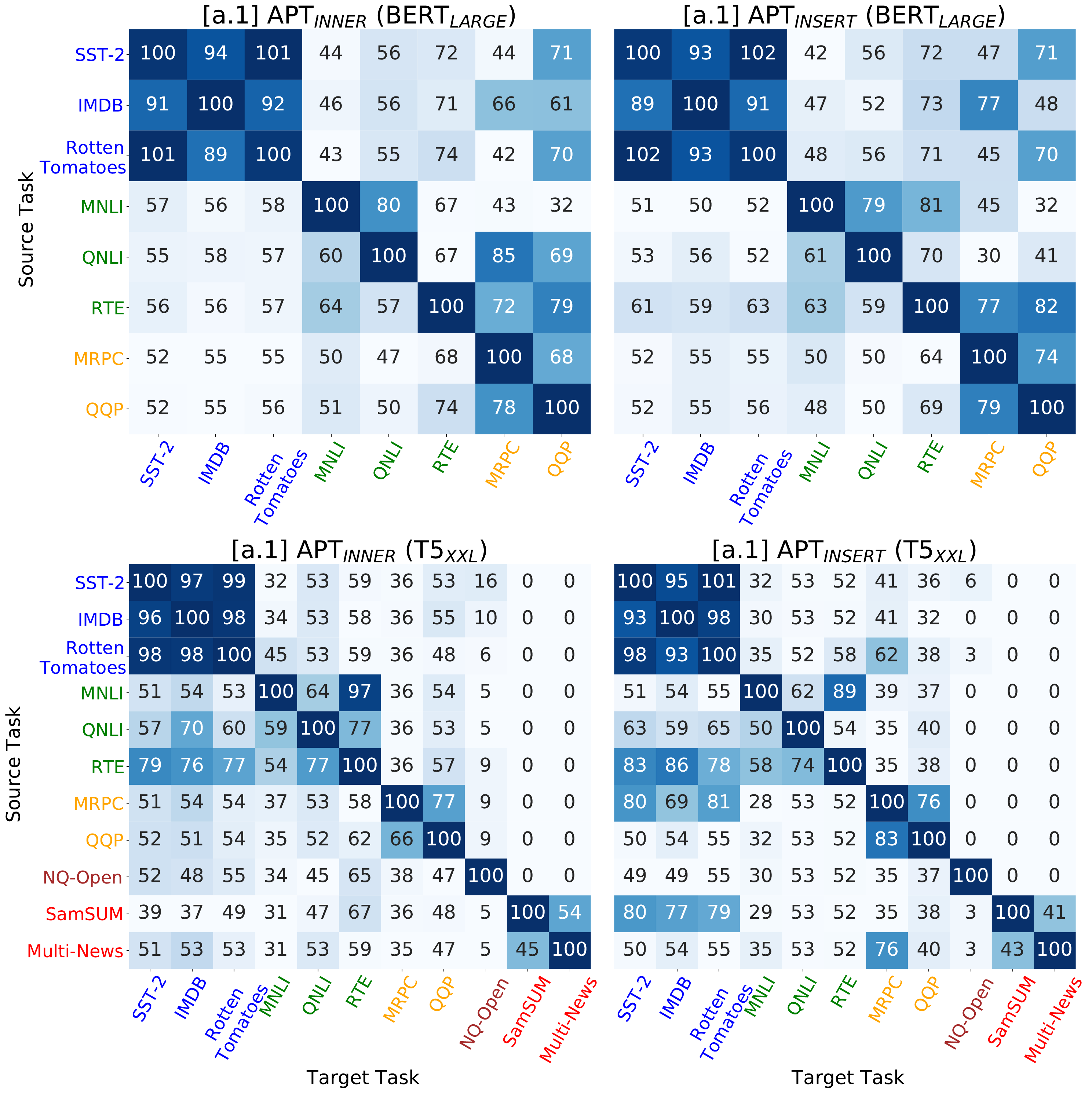}
    }
    \caption{Relative performance (zero-shot transfer performance / original performance) (\%) on the target tasks (columns) of the \APET methods trained on the source tasks (rows). Colors of the task names indicate task types.}
    \label{fig:task_transfer}
\end{figure*}

\subsection{Training Configurations of \APET methods}
\label{appendix:training_configurations_of_afp_methods}
The tunable module of a \APET method $\theta$ is composed of $L$ tunable weights $\mathbf{W}$, which can be expressed as $\theta= \{ \mathbf{W}_{1}, \mathbf{W}_{2}, ..., \mathbf{W}_{L} \}$. We also follow Equation \eqref{eq:pet_training_objective} to train the \APET method. During training, we only optimize $\theta$ while freezing the rest of the parameters in the PLM. 

Besides, we adopt a batch size of $32$ and have no warm-up for most of the \APET models and tasks. In addition, The maximum input length is $128$ for single sentence tasks (SA) and $256$ for multi-sentence tasks (NLI, PI, QA, SUM). The maximum generation length is $1$ for classification tasks (SA, NLI, PI), $64$ for Multi-News, and $128$ for SAMSum. On the BERT, BLOOM, T5 models, we set their learning rates as \{3e-4\}, \{3e-4, 5e-5\}, \{1e-4, 1e-3, 1e-2\} respectively. Then, we choose the best performance to report.

\section{Number of Tunable Parameters of \APET}
\label{appendix:apet_parameter_number}

Here, the Table \ref{table:apet_parameter_number} shows the number of tunable parameters of \APET for each group in \figref{fig:structure_and_position}.

\begin{table}[!t]
\begin{center} 
\scalebox{0.675}
{{
\begin{tabular}{c|cccc}
\toprule
\multicolumn{1}{c|}{} & \multicolumn{1}{c}{$\text{APET}_{\text{Prompt}}$} & \multicolumn{1}{c}{$\text{APET}_{\text{BitFit}}$} & \multicolumn{1}{c}{$\text{APET}_{\text{LoRA}}$}  & \multicolumn{1}{c}{$\text{APET}_{\text{Adapter}}$} \\ 
\midrule
{\BERTSMALL} & {5.1e+4} & {1.4e+4} & {6.6e+4} & {2.0e+5} \\
\midrule
{\BERTLARGE} & {1.0e+5} & {1.7e+5} & {7.9e+5} & {2.4e+6} \\
\midrule
\midrule
{\BLOOMSMALL} & {1.0e+5} & {2.7e+5} & {7.9e+5} & {2.4e+6} \\
\midrule
{\BLOOMLARGE} & { 4.1e+5} & {1.4e+6} & { 3.9e+6} & {1.2e+7} \\
\midrule
\midrule
{\TSMALL} & { 5.1e+4} & {1.2e+5} & {2.3e+5} & {8.0e+5} \\
\midrule
{\TXXL} & {4.1e+5} & {2.5e+6} & {6.3e+6} & {1.9e+7} \\
\bottomrule
\end{tabular}
}}
\caption{The table shows the numbers of tunable parameters of tunable parameters in \figref{fig:structure_and_position} experiment.}
\label{table:apet_parameter_number}
\end{center} 
\end{table}

\section{Power of Model Scale to Transferability}
\label{sec:the_power_of_model_scale_to_transferability}
Furthermore, to explore whether the power of model scale can also facilitate generalization ability of tuning methods, we explore the transferability between the NLP tasks in the zero-shot setting \cite{vu-etal-2022-spot,transferabilitysu,deltatuning}. In the experiments, we first train the parameters of \APET methods on the source tasks and directly reuse them on the target tasks in zero-shot setting. We will investigate two series of PLMs T5 (\TSMALL and \TXXL) and BERT (\BERTSMALL and \BERTLARGE) and report the relative performance.

Note that for different types of tasks, they are expected to share different groups of label sets (e.g. for task like SA, the labels are usually positive/negative, whereas, for tasks like NLI, the labels are usually entailment/not entailment). Reusing the parameters trained on the source task to test on the target task will naturally fail since the model is not able to generate the labels they have never seen in the training stage. To this end, we generally map the original label sets to a unified label set (e.g. negative/not entailment/false --> 0, positive/entailment/true --> 1). Utilizing a unified label set makes it feasible to evaluate the transferability of the AFP method among different types of tasks regardless of the divergence of original labels.

The results are shown in \figref{fig:task_transfer}, from which we can find that the \APET (\APETDISCRETE and \APETADJACENT) methods can transfer to the same type of tasks demonstrated by the darker color alongside the diagonal of the matrix and generally perform well both on small-scale PLMs (\figref{fig:task_transfer} (a): \BERTSMALL and \TSMALL) and large-scale PLMs (\figref{fig:task_transfer} (b): \BERTLARGE and \TXXL). However, the lighter color indicates that \APET methods have difficulty performing different types of tasks overall, and both small-scale and large-scale PLMs share this phenomenon. This finding indicates that the power of scale does not necessarily facilitate the generalization ability of AFP methods which is in line with the prevalent assumption that fewer parameters often cause underfitting, whereas more parameters tend to cause overfitting. Nevertheless, the mechanism behind this phenomenon still arouses our deep concern and is worth expanding that we will systematically analyze it in our future work.



\end{document}